\documentclass{article}

\usepackage[preprint]{neurips_2019}

\usepackage[utf8]{inputenc} % allow utf-8 input
\usepackage[T1]{fontenc}    % use 8-bit T1 fonts
\usepackage{hyperref}       % hyperlinks
\usepackage{url}            % simple URL typesetting
\usepackage{booktabs}       % professional-quality tables
\usepackage{amsfonts}       % blackboard math symbols
\usepackage{nicefrac}       % compact symbols for 1/2, etc.
\usepackage{microtype}      % microtypography

\usepackage{xcolor,bm,amsmath,graphicx,comment,bigints}
\usepackage{algorithm,algorithmic}
\graphicspath{{images/}}

\definecolor{plum}{rgb}{0.45,0,.66}
\definecolor{mythistle}{rgb}{.99,.195,.133}
\definecolor{myred}{cmyk}{0.000000,1.000000,1.000000,0.1}
\definecolor{myblue}{cmyk}{1.000000,0.750000,0.000000,0.1}
\definecolor{mybgn}{cmyk}{0.850000,0.350000,0.000000,0.1}
\definecolor{mygrn}{cmyk}{0.750000,0.000000,1.000000,0.2}
\newcommand{\bl}{\begin{itemize}}
\newcommand{\el}{\end{itemize}}
\newcommand{\be}{\begin{enumerate}}
\newcommand{\ee}{\end{enumerate}}

\newcommand{\bea}{\begin{eqnarray*}}
\newcommand{\eea}{\end{eqnarray*}}

\newcommand{\beq}{\begin{equation}}
\newcommand{\eeq}{\end{equation}}

\newcommand{\bmx}{\left[ \begin{array}}
\newcommand{\emx}{\end{array} \right]}

\newcommand{\mybf}[1]{\textbf{\em #1} }

\def\bfx{{\mybf{x}}}

\DeclareMathOperator{\erf}{erf}
\DeclareMathOperator{\softmax}{softmax}

\definecolor{mybrick}{rgb}{0.82, 0.1, 0.26}

\newcommand{\achange}{}
\newcommand{\bchange}{}
\newcommand{\cchange}{}

\title{Investigating Decision Boundaries of Trained Neural Networks}

\author{%
  Roozbeh Yousefzadeh \\%\thanks{Use footnote for providing further information about author (webpage, alternative address)---\emph{not} for acknowledging funding agencies.} \\
  Department of Computer Science \\
  University of Maryland \\
  College Park, MD 20742 \\
  \texttt{roozbeh@cs.umd.edu} \\
  \And
  Dianne P. O'Leary \\
  Department of Computer Science and\\
  Institute for Advanced Computer Studies \\
  University of Maryland \\
  College Park, MD 20742 \\
  \texttt{oleary@cs.umd.edu} \\
}

\begin{document}

\maketitle

\begin{abstract}
Deep learning models have been the subject of study from various perspectives, for example, their training process, interpretation, generalization error, robustness to adversarial attacks, etc. A trained model is defined by its decision boundaries, and therefore, many of the studies about deep learning models speculate about the decision boundaries, and sometimes make simplifying assumptions about them. So far, finding exact points on the decision boundaries of trained deep models has been considered an intractable problem. Here, we compute exact points on the decision boundaries of these models and provide mathematical tools to investigate the surfaces that define the decision boundaries. Through numerical results, we confirm that some of the speculations about the decision boundaries are accurate, some of the computational methods can be improved, and some of the simplifying assumptions may be unreliable, for models with nonlinear activation functions. We advocate for verification of simplifying assumptions and approximation methods, wherever they are used. Finally, we demonstrate that the computational practices used for finding adversarial examples can be improved and computing the closest point on the decision boundary reveals the weakest vulnerability of a model against adversarial attack.
\end{abstract}

\section{Introduction}
Interpreting the behavior of trained neural networks, their generalization error, and robustness to adversarial attacks are open research problems that all deal, directly or indirectly with the decision boundaries of these models. %For interpretation of trained models, it is important to find the least changes that would flip the output of the model, commonly referred to as local interpretation; and to find the properties of subdomains in the input space that produce the same output, referred to as global interpretation. 
The decision boundaries of neural networks have typically been investigated through simplifying assumptions or approximation methods. As we will show in our numerical results, many of these simplifications may lead to unreliable results.

In this work, we show that better information about interpretation, generalization error, and robustness of a model can be obtained by computing {\em flip points}. Flip points are points on the decision boundaries of the model. For any data point, its  \underline{closest} flip point is the closest point that would flip the decision of the model to indeterminate. 
The direction between the data point and its closest flip point reveals a great deal about the
most influential features in decision making of the model. %\citep{yousefzadeh2019interpreting}. 

Some previous work has highlighted the importance of boundary points.
Studies such as \cite{lippmann1987anintroduction}  investigate the decision boundaries of single-layer perceptrons, while describing the difficulties that arise regarding the complexity of decision boundaries for multi-layer networks.
While \cite{spangher2018actionable}  proposed a method to find the least changes in the input that would flip the classification of a model, their method is applicable only to linear classifiers, and they do not actually investigate the decision boundary. 
\cite{wachter2018counterfactual} defined \textit{counterfactuals} as the possible changes in the input that can produce a different output label. But, for a continuous model, the closest counterfactual is ill-defined, since there are points arbitrarily close to the decision boundaries that can produce different output labels. Furthermore, they solve their problem by enumeration, applicable only to a small number of features.

Some studies on interpretation of deep models have made simplifying assumptions about the decision boundaries. For example, \cite{ribeiro2016should} assumes that the decision boundaries are locally linear. Their approach tries to sample points on two sides of a decision boundary, then perform a linear regression to approximate the decision boundary  and explain the behavior of the model. However, as we show in our numerical results, decision boundaries of  neural networks  can be highly nonlinear, even locally, and a linear regression can lead to unreliable explanations. %\cite{anchors_aaai18} also considers the decision boundaries to be linear and describes subdomains in the input space that produce the same output, by approximating the subdomains with .

Regarding the generalization error of trained models, \cite{elsayed2018large} and \cite{marginbased2019} have shown there is a relationship between the closeness of training points to the decision boundaries and the generalization error of a model. However, they regard computing the distance to the decision boundary as an intractable problem and instead use the derivatives of the output to derive an approximation to the closest distance. In our numerical results, we compare their approximation to our results, and show the advantages of computing the distance directly. 
Other studies such as \cite{neyshabur2017exploring} approximate the closest distance to the decision boundary by the closest distance to an input with another label, which can be an overestimate. They use the modified margin between $\softmax$ outputs as a measure of distance, which we see later can be misleading.
%We further show that investigating flip points for correct and wrong classifications of a network on a validation set can reveal patterns that cause the mistakes. 
%We use these patterns to adjust the mistakes of the model, which leads to reducing the generalization error.

Regarding adversarial attacks, there are many studies that seek small perturbations in an input that can change the classification of the model. \cchange{For example, \cite{moosavi2016deepfool,jetley2018friends,fawzi2017robustness} apply small perturbations to the input until its classification changes, but, since they do not attempt to find the closest point on the decision boundaries of the model, they do not reveal its weakest vulnerabilities.} \bchange{Most recent studies on adversarial examples, such as} \cite{tsipras2018robustness} and \cite{ilyas2019adversarial} minimize the loss function of the neural network for the adversarial label, subject to a distance constraint. They impose the distance constraint in order to find an adversarial example similar to the original image. Although this method is an important tool, this form of seeking adversarial examples has certain limitations, regarding the ability to make the models robust, and regarding the measurement of robustness of models, as we explain through numerical examples. We show that finding the closest point on the decision boundary accurately represents the least perturbation needed for adversarial classification, and, therefore, studies on adversarial examples can benefit from direct investigation of decision boundaries.

\section{Computing the closest flip point}

In the results we show later, the output of the neural network has been computed using $\softmax$, which gives us a convenient normalization. For a given data point and its output class, we compute the closest flip point with respect to another output class by solving a numerical optimization problem: minimizing the distance between the given data point and the unknown (flip) point, subject to the constraints that, at the unknown point, the $\softmax$ outputs  for the two classes are equal, and that no other $\softmax$ output is greater \citep{yousefzadeh2019interpreting}.

Our optimization problem is nonconvex, so we cannot be sure that optimization algorithms will find the global minimizer. One important fact that makes the optimization easier is that we have a good starting point, the data point itself. We have solved our optimization problem using the applicable algorithms in 3 packages, NLopt \citep{nlopt}, IPOPT \citep{wachter2006implementation}, and Optimization Toolbox of MATLAB, as well as our own custom-designed homotopy algorithm. The algorithms almost always converge to the same point; in fewer than 5\% of images, the interior point algorithms find closer flip points. \achange{The variety and abundance of global and local optimization algorithms in the above optimization packages give} us confidence that we have indeed usually found the closest flip point. In any case, we demonstrate below that our flip points are closer than those estimated by methods such as linear approximations.

In our numerical results on image data, we measure  distance  using the $\ell_2$ norm. In general, the choice of distance measure should be guided by practitioners who understand the nature of the data.

\section{Numerical experiment setup}

To illustrate our ideas, we use a 12-layer feedforward neural network trained on 2 classes of the CIFAR-10 dataset, ships and planes; details are provided in an appendix. To train the network, we have used Tensorflow, with Adam optimizer, learning rate of 0.001, and Dropout with rate 50\%.

We use a tunable error function as the activation function. This allows us to introduce nonlinearity into the model while having control over the magnitude of the derivatives. Keep in mind that one can compute flip points for trained models and interpret them regardless of the architecture of the model (number of layers, activation function, etc.), the training set, and the training regime (regularization, etc.). 

\begin{figure}[h]
\centering
  \begin{minipage}[b]{0.19\columnwidth}
\centerline{\includegraphics[width=1\columnwidth]{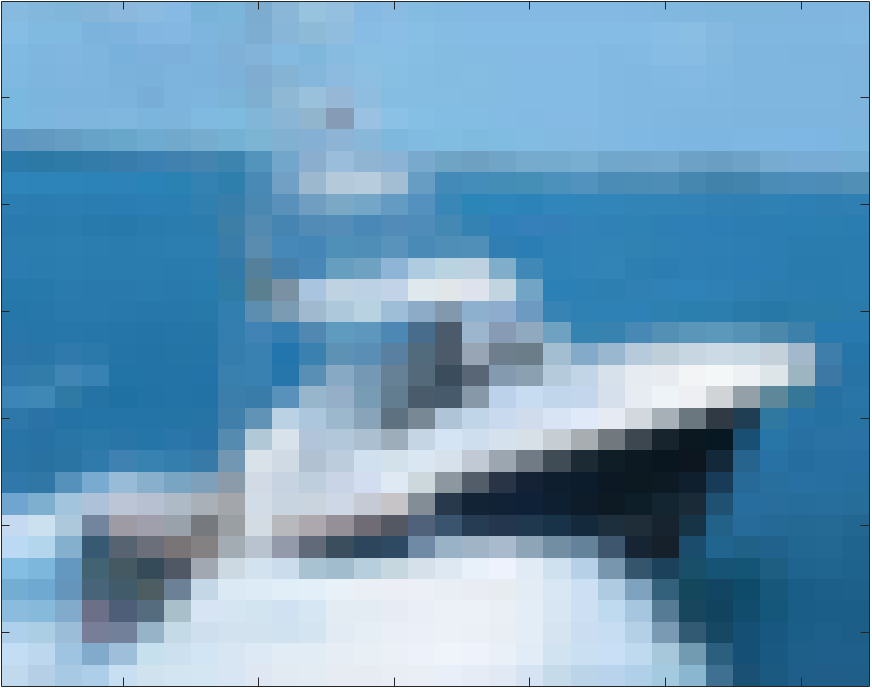}}
  \end{minipage}
  \begin{minipage}[b]{0.19\columnwidth}
\centerline{\includegraphics[width=1\columnwidth]{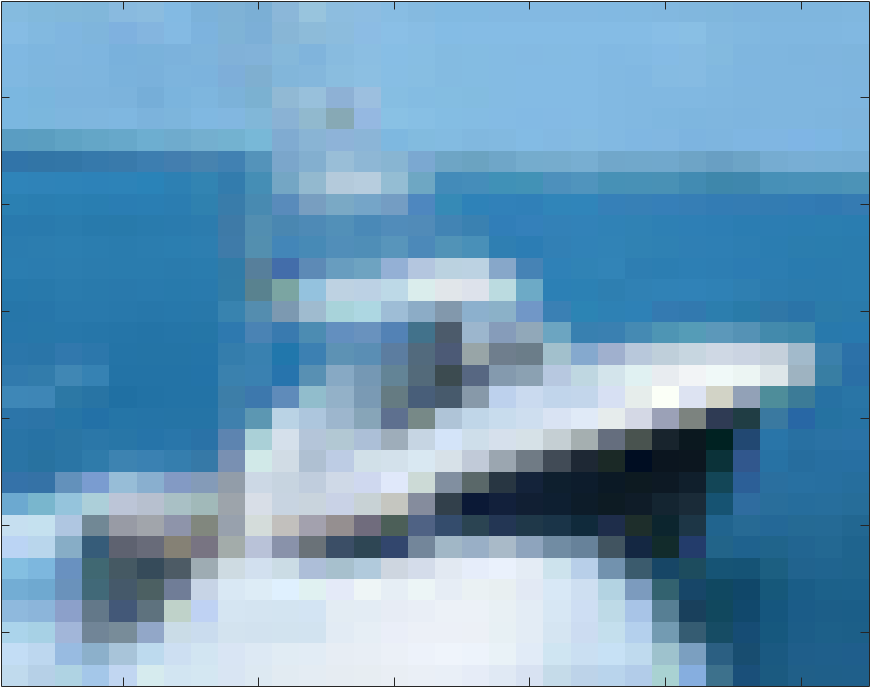}}
  \end{minipage}
  \begin{minipage}[b]{0.19\columnwidth}
\centerline{\includegraphics[width=1\columnwidth]{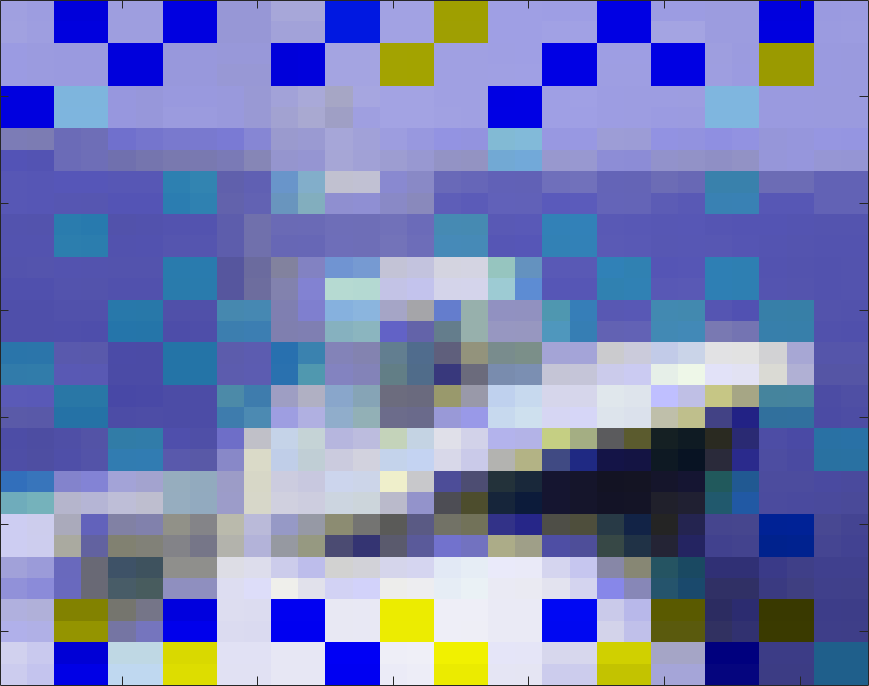}}
  \end{minipage}
  \begin{minipage}[b]{0.19\columnwidth}
\centerline{\includegraphics[width=1\columnwidth]{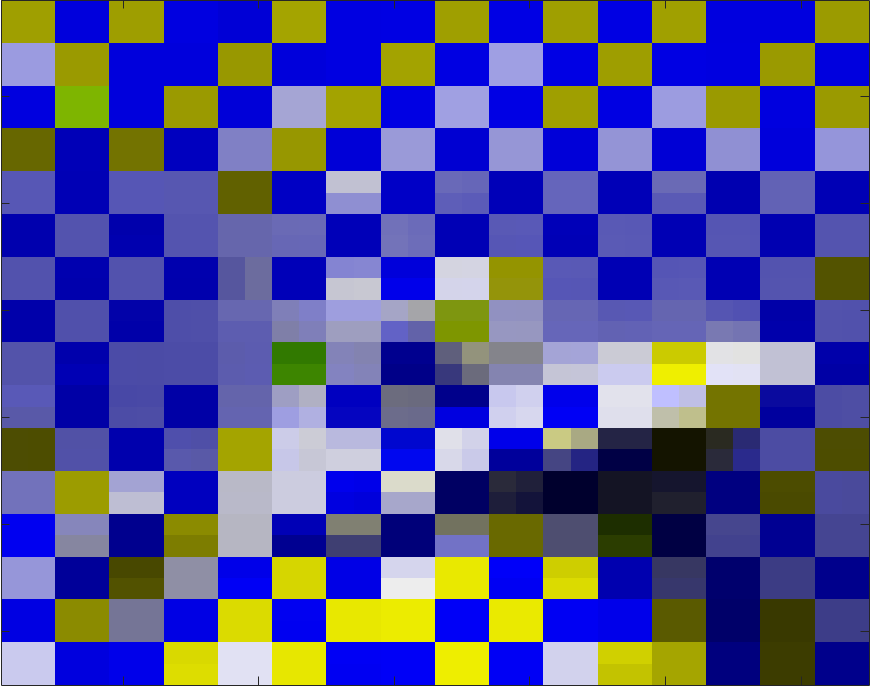}}
  \end{minipage}
  \begin{minipage}[b]{0.19\columnwidth}
\centerline{\includegraphics[width=1\columnwidth]{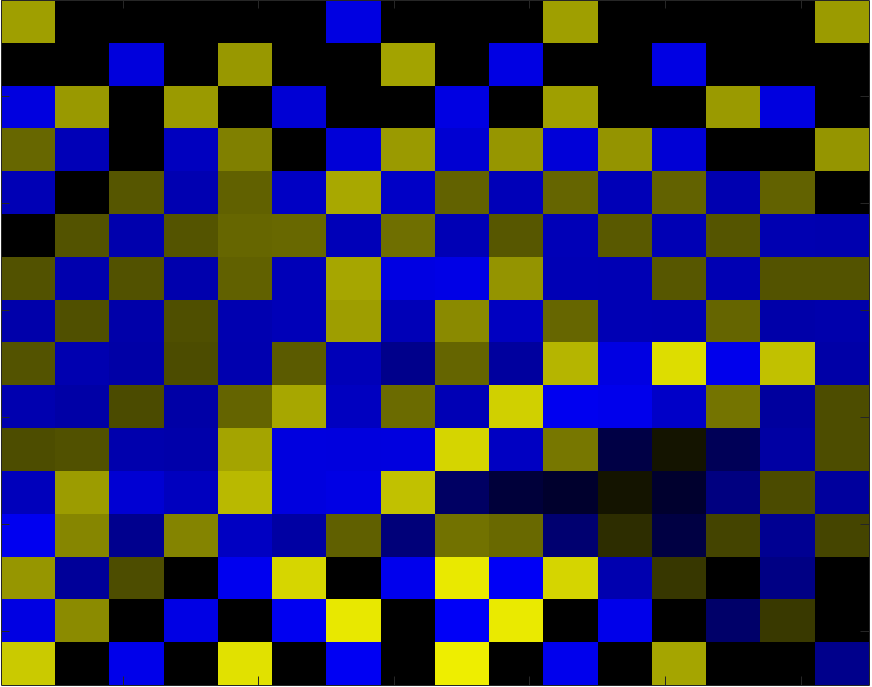}}
  \end{minipage}
\caption{Reconstruction of an image from a subset of wavelet coefficients leads to different representations. The original image (left), with 4096 wavelet coefficients, is reconstructed using the  most significant 2200, 1000, 500, and 200 wavelet coefficients (respectively, from left to right), chosen according to pivoted QR factorization.}
\label{fig_reconstruct}
\end{figure}

Inputs to our network are not  the pixels, but 200 of the 3D Daubechies-1 wavelet coefficients. We choose the 200 coefficients according to the pivoted QR factorization \citep{golub2012matrix} of the wavelet coefficients for the training set.
Using the most significant wavelet coefficients removes  redundancies in the features of the image. %It also allows us to train small networks that are easier to train and to work with. 
Figure \ref{fig_reconstruct} shows the first ship image in the CIFAR-10 training set along with its reconstructions from subsets of wavelet coefficients. With fewer coefficients, the reconstructed image looks less similar to the original image. Nevertheless, the model is able to correctly classify most of the images by learning from those representations. This result may be in agreement with the arguments of \cite{ilyas2019adversarial} that neural networks learn Gaussian representations of images. 

The accuracy we obtain on the testing set is 84.05\%. This can be improved to near 95\% using the calculated flip points as new training points in order to move decision boundaries, or by using more wavelet coefficients. \achange{Notice that using a smaller number of wavelet coefficients leads to a model with smaller feature space and that makes us more likely to find the global solution of our non-convex optimization problem, or a solution close to it.}

In our computations, we verify that each computed flip point is a legitimate image, satisfying appropriate upper and lower bounds for each pixel.
We never encountered a case in which these bounds were violated, but violations could be handled by extra constraints or by projection.

\section{Investigating the neural network function and the closest flip points}
Here, viewing  the trained neural network as a function, we investigate the paths between inputs and  flip points.
% and the optimality of flip points we find on the decision boundaries.

\subsection{Lipschitz continuity of the output of trained model}

The output of our neural network is a smooth mathematical function. Because it is the composition of a finite set of Lipschitz continuous functions, the output is also Lipschitz continuous. The Lipschitz constant can be bounded using the tuning constants for the error functions and the norms of the matrices applied at each layer.
%In Appendix A, we investigate the Lipschitz continuity of our neural network function and estimate the Lipschitz constant for it. Our estimation of Lipschitz constant is based on two separate approaches. First approach is a formal conservative approach based on the norms of matrices in the chain rule decomposition of derivatives. Second approach is based on numerical maximization of the derivatives in the input space. The results we obtain from the two approaches support each other.

Why does this matter? As we walk along a direct path connecting one data point to another, the Lipschitz constant can tell us \achange{how fine we should discretize that path in order to accurately depict the output of network and identify the locations of decision boundaries. This means that we choose the distance between the discretization points small enough such that the output of network can be considered to change linearly between any consecutive points, with negligible error.% The Lipschitz constant for our trained network is upper bounded by $4.7\times10^{-7}$.
}

%big a step we can take and still be sure that the classification does not change. Thus we can be assured that we do not inadvertently step across a decision boundary without noticing it.

%Having an estimate of the Lipschitz constant allows us to have an estimate of how much the output of the model may change, when inputs are changed. Therefore, if we choose two points close enough in the input space, and evaluate the output of network for them, a line may be able to approximate the output of the network between those points, with bounded error. This bounded error on the $\softmax$ score can ensure that the classification of the model does not change between two points that are close enough.

\subsection{\achange{Investigating output of network along direct paths between images}}

Here, we draw lines between images, discretize those lines, and plot the output of the network along them.
Consider two images, $\bfx_1$ and $\bfx_2$, separated by distance $d= \| \bfx_1 - \bfx_2 \|_2$. The points on the line connecting them may be defined by $(1 - \alpha) \bfx_1 + \alpha \bfx_2$ where $\alpha$ is a scalar between 0 and 1. This line can be extended beyond $\bfx_1$ and $\bfx_2$ on either side by choosing $\alpha < 0$ or $\alpha >  1$, respectively.

\begin{figure}[h]
\centering
  \begin{minipage}[b]{0.49\columnwidth}
\centerline{\includegraphics[width=1\columnwidth]{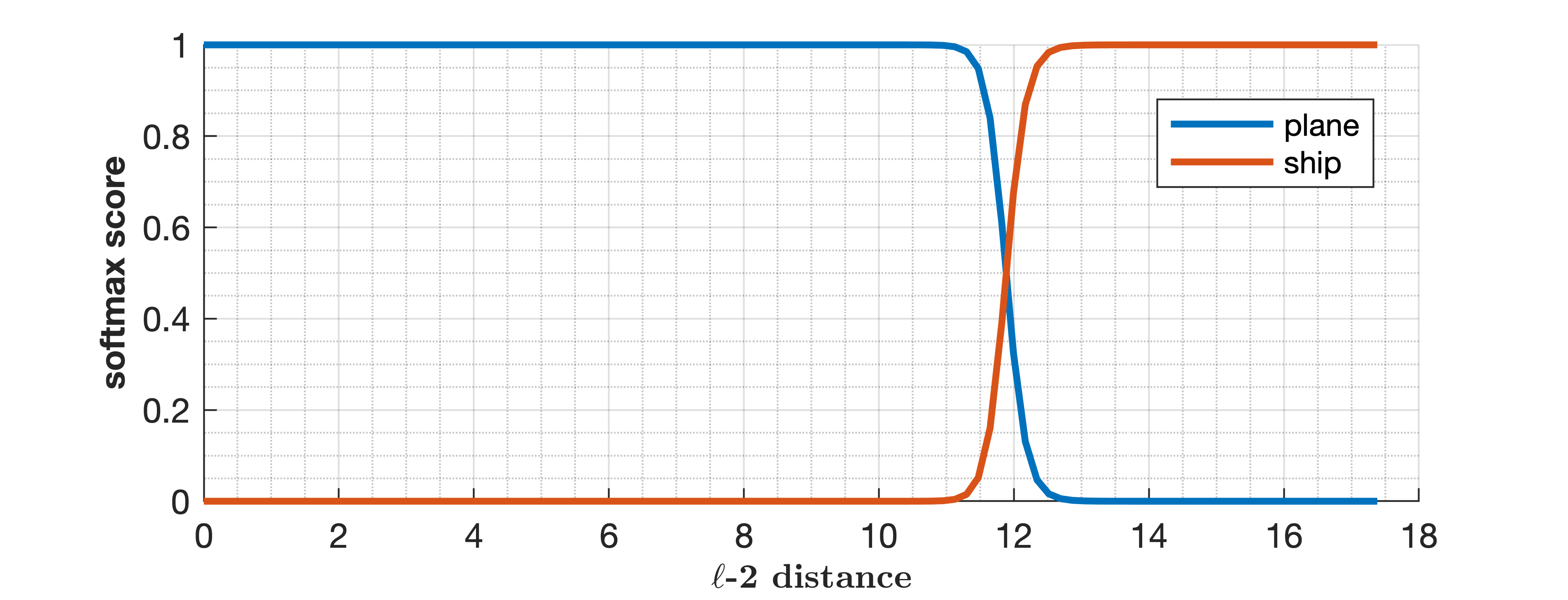} }
  \end{minipage}
  \begin{minipage}[b]{0.49\columnwidth}
\centerline{\includegraphics[width=1\columnwidth]{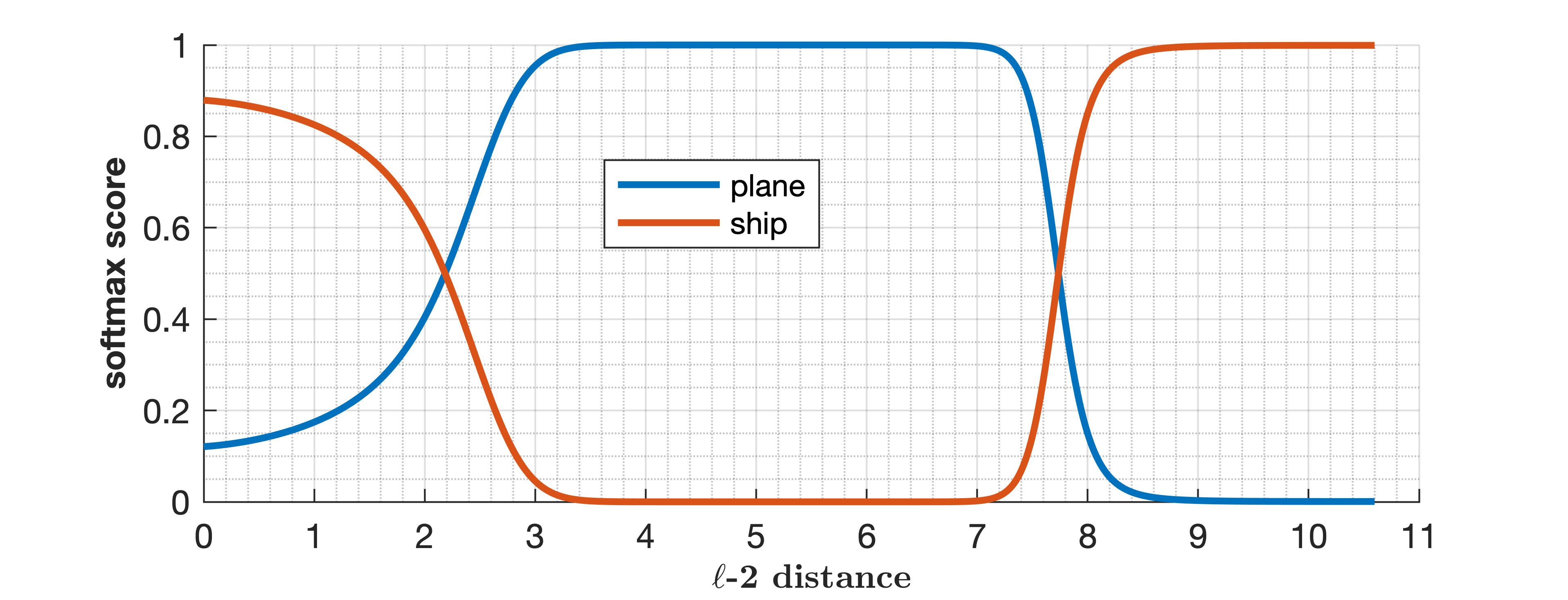} }
  \end{minipage}
  \begin{minipage}[b]{0.49\columnwidth}
\centerline{\includegraphics[width=1\columnwidth]{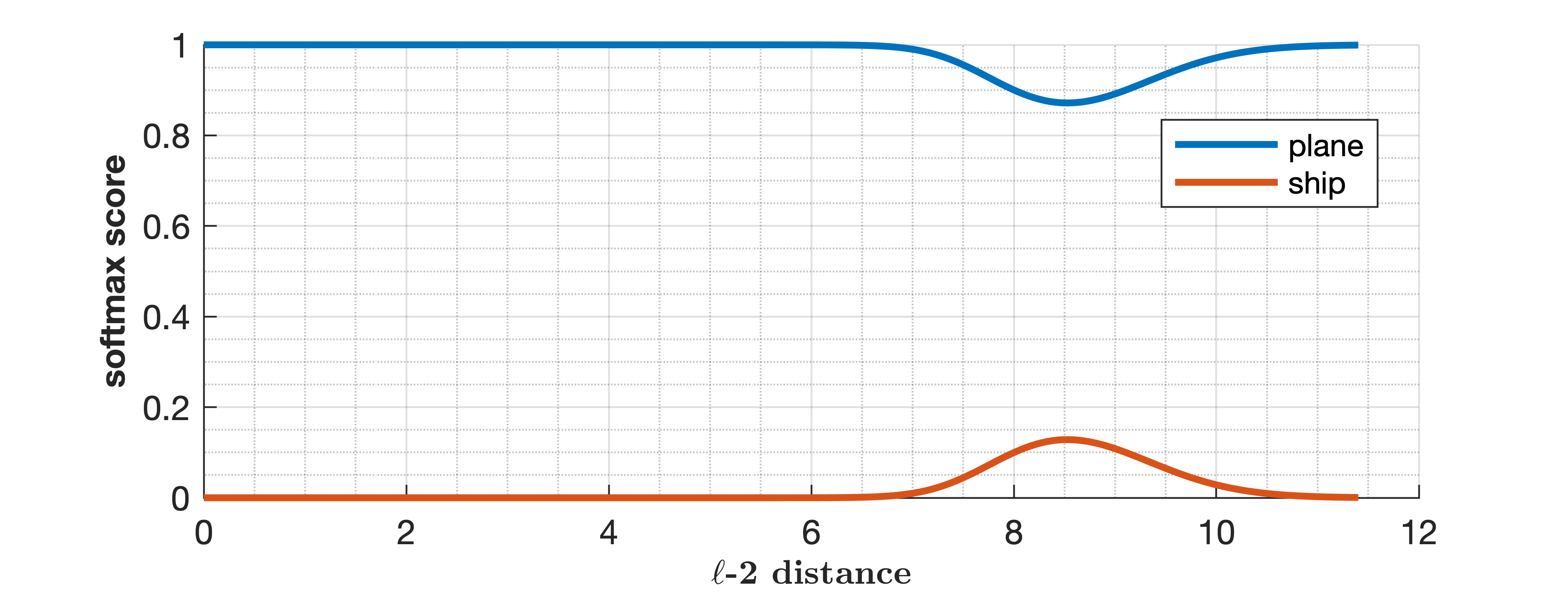} }
  \end{minipage}
  \begin{minipage}[b]{0.49\columnwidth}
\centerline{\includegraphics[width=1\columnwidth]{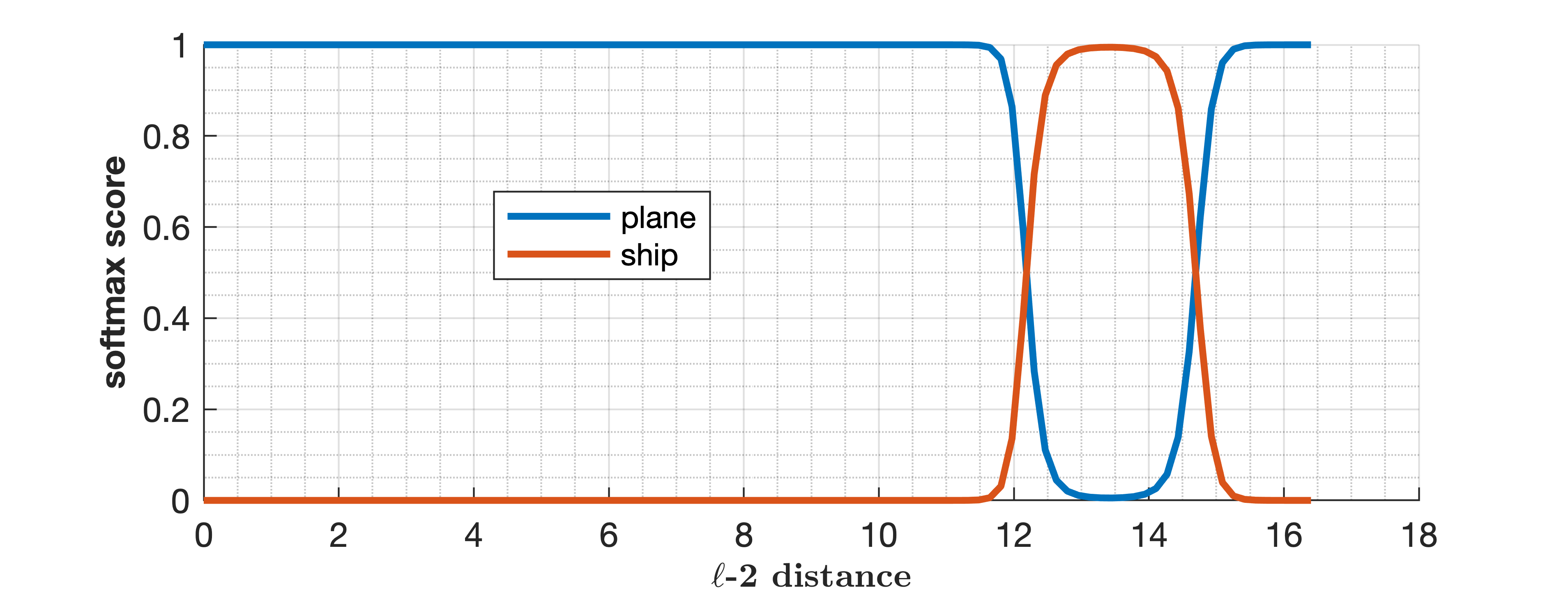} }
  \end{minipage}
  \begin{minipage}[b]{0.49\columnwidth}
\centerline{\includegraphics[width=1\columnwidth]{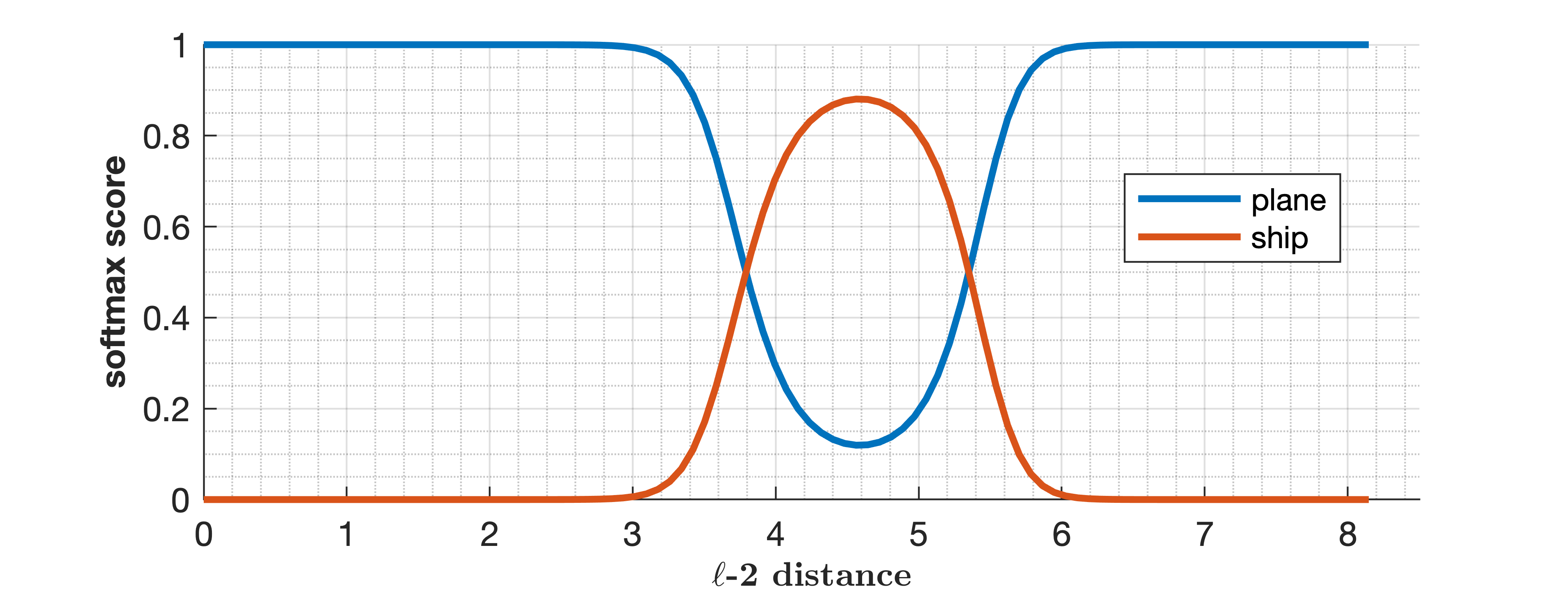} }
  \end{minipage}
  \begin{minipage}[b]{0.49\columnwidth}
\centerline{\includegraphics[width=1\columnwidth]{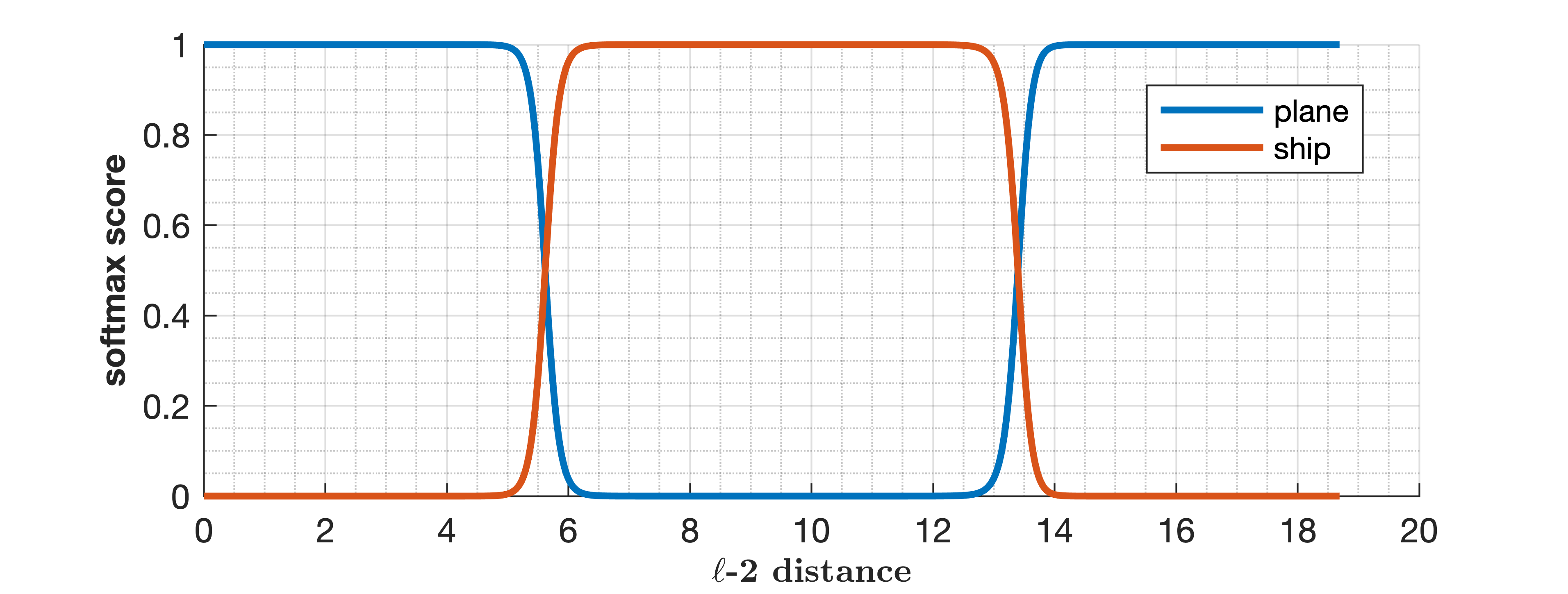} }
  \end{minipage}
\caption{Model output along the line connecting two images. }
\label{fig_ray1}
\end{figure}

In Figure \ref{fig_ray1}, zero on the horizontal axis corresponds to image $\bfx_1$ and the right boundary corresponds to $\bfx_2$, an image chosen from the same or other class.
%Either of them can be images in the training or testing sets, or any of the flip points. 
%When choosing $\alpha > 1$, we are extending the line beyond $\bfx_2$ and therefore we would have points on the x-axis farther than $d$. When choosing $\alpha < 0$, we are extending the line beyond $\bfx_1$, in opposite direction from $\bfx_2$, hence, we negate the distance. In this setting, $\alpha = -1$ will correspond to $-d$ on the x-axis.
%As explained in Appendix A, the fineness of discretization is chosen according to the estimate of Lipschitz constant, in order to ensure any visible fluctuation of the $\softmax$ score gets captured in our plots. 
The lines connecting most pairs of images in the data set resemble the top left plot in Figure \ref{fig_ray1} in their simplicity; both images are far from the decision boundary, and the line between them crosses the decision boundary once. 
The other five plots in this figure are hand picked to demonstrate atypical cases. Having multiple boundary crossings is more frequent among the images in the testing set, compared to images in the training set. 
%This implies that the output of network might not be well-defined between inputs, which is interesting to study further. 

\begin{figure}[h]
\centering
  \begin{minipage}[b]{0.49\columnwidth}
\centerline{\includegraphics[width=1\columnwidth]{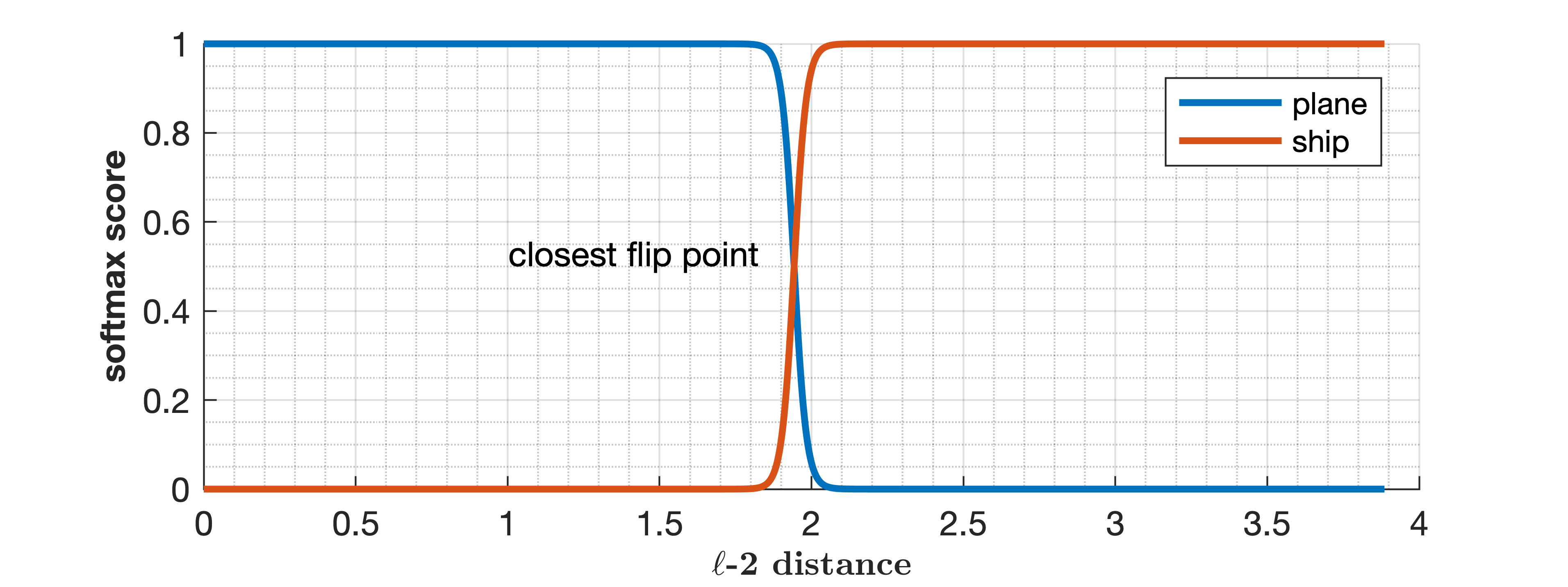} }
  \end{minipage}
  \begin{minipage}[b]{0.49\columnwidth}
\centerline{\includegraphics[width=1\columnwidth]{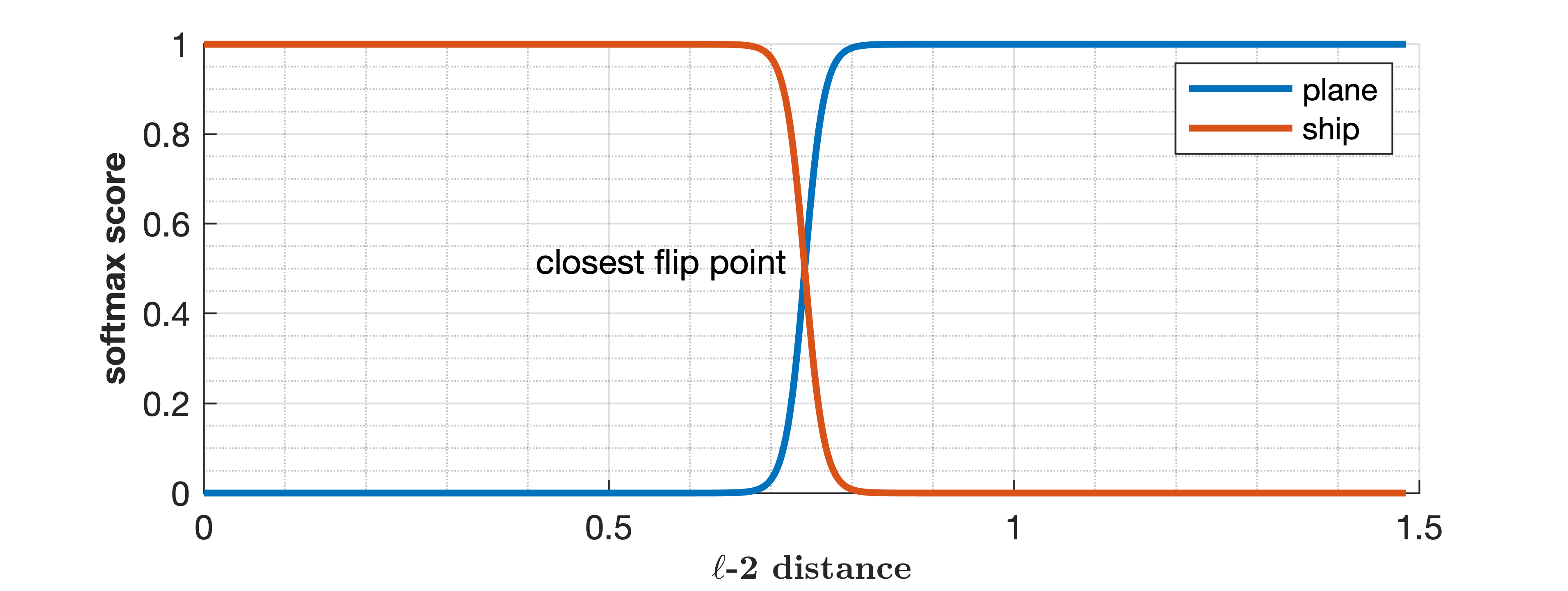} }
  \end{minipage}
  \begin{minipage}[b]{0.49\columnwidth}
\centerline{\includegraphics[width=1\columnwidth]{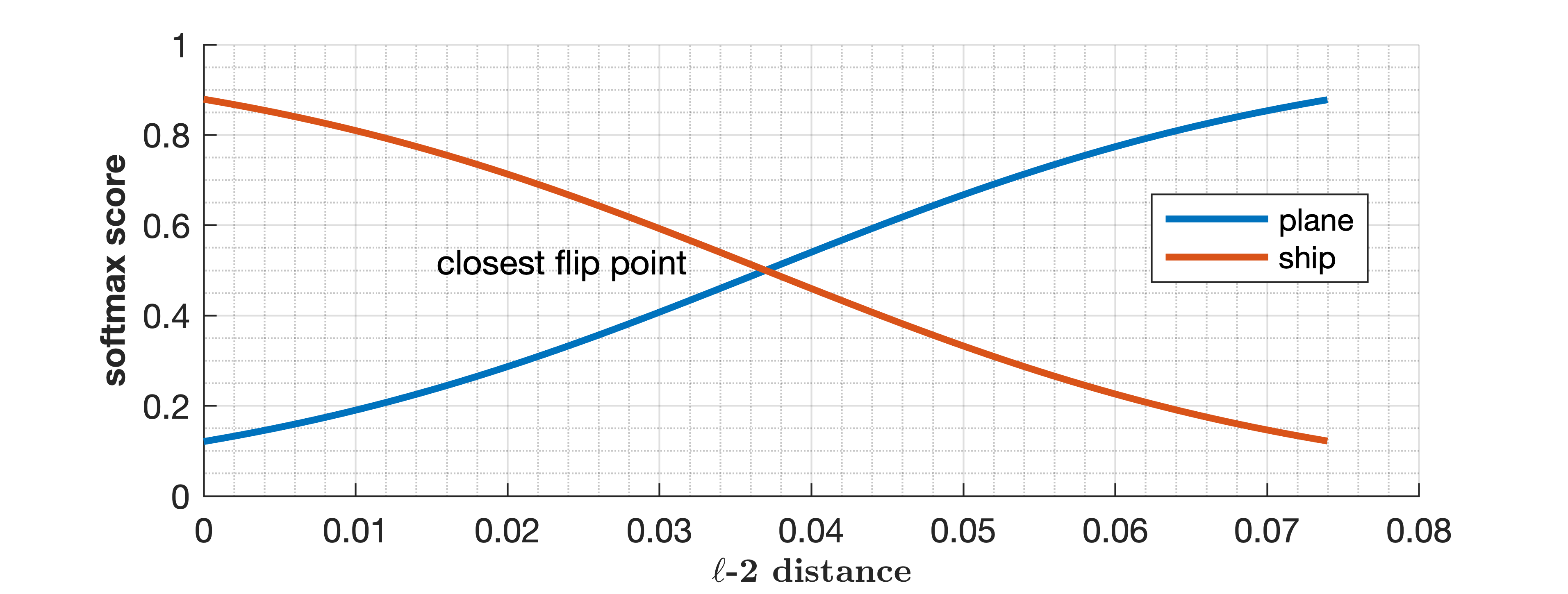} }
  \end{minipage}
  \begin{minipage}[b]{0.49\columnwidth}
\centerline{\includegraphics[width=1\columnwidth]{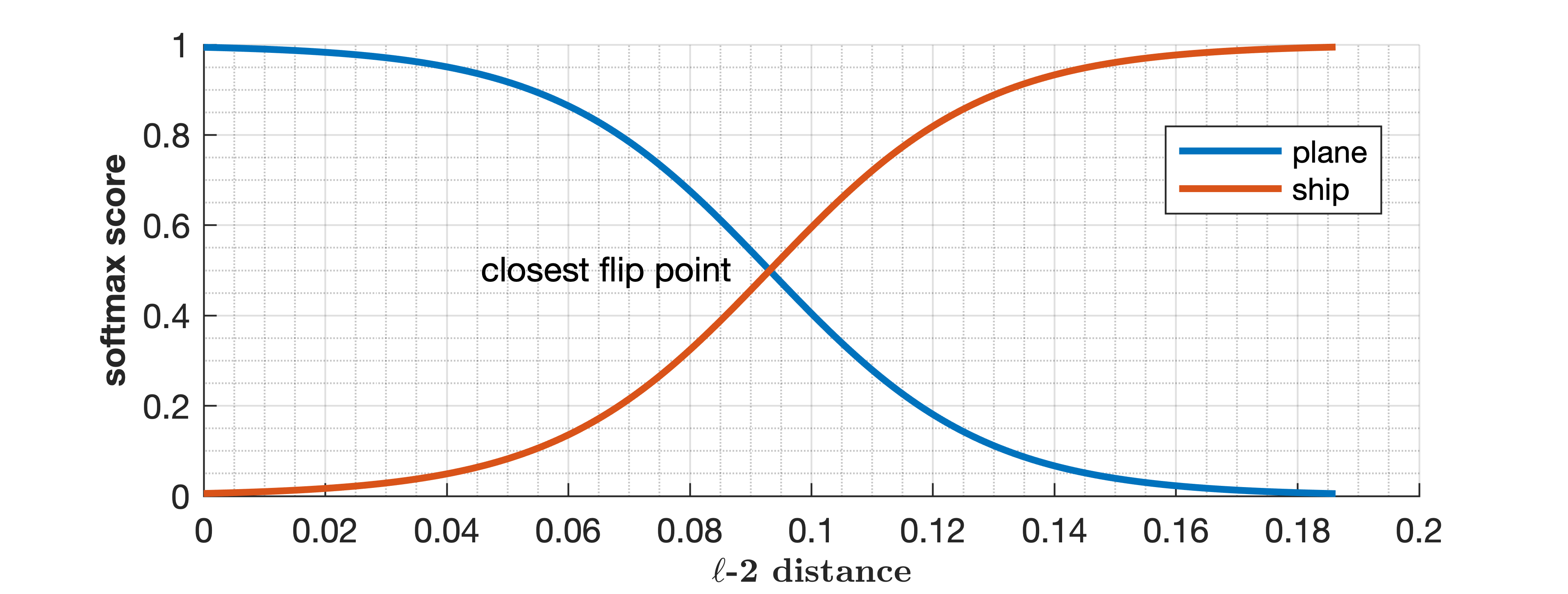} }
  \end{minipage}
\caption{Model output along the line connecting an image with its closest flip point. Images for the top row  are correctly classified, while images for  the bottom row  are misclassified. }
\label{fig_ray2}
\end{figure}

Figure \ref{fig_ray2}  shows the output of the model for some lines connecting images to their closest flip points. Notice that the two bottom plots have a much smaller distance scale, and the behavior of the $\softmax$ score for correctly and incorrectly classified points is quite similar. \achange{These plots clearly show that the decision boundaries \bchange{in our model} are far from linear and a hyperplane would not be able to approximate such boundary surfaces. \cite{fawzi2018empirical} also have the view that the decision boundaries of deep models are highly curved, but they had not computed exact points on the decision boundaries. Our results confirm their view.}

\begin{figure}[h]
\centering
  \begin{minipage}[b]{0.49\columnwidth}
\centerline{\includegraphics[width=1\columnwidth]{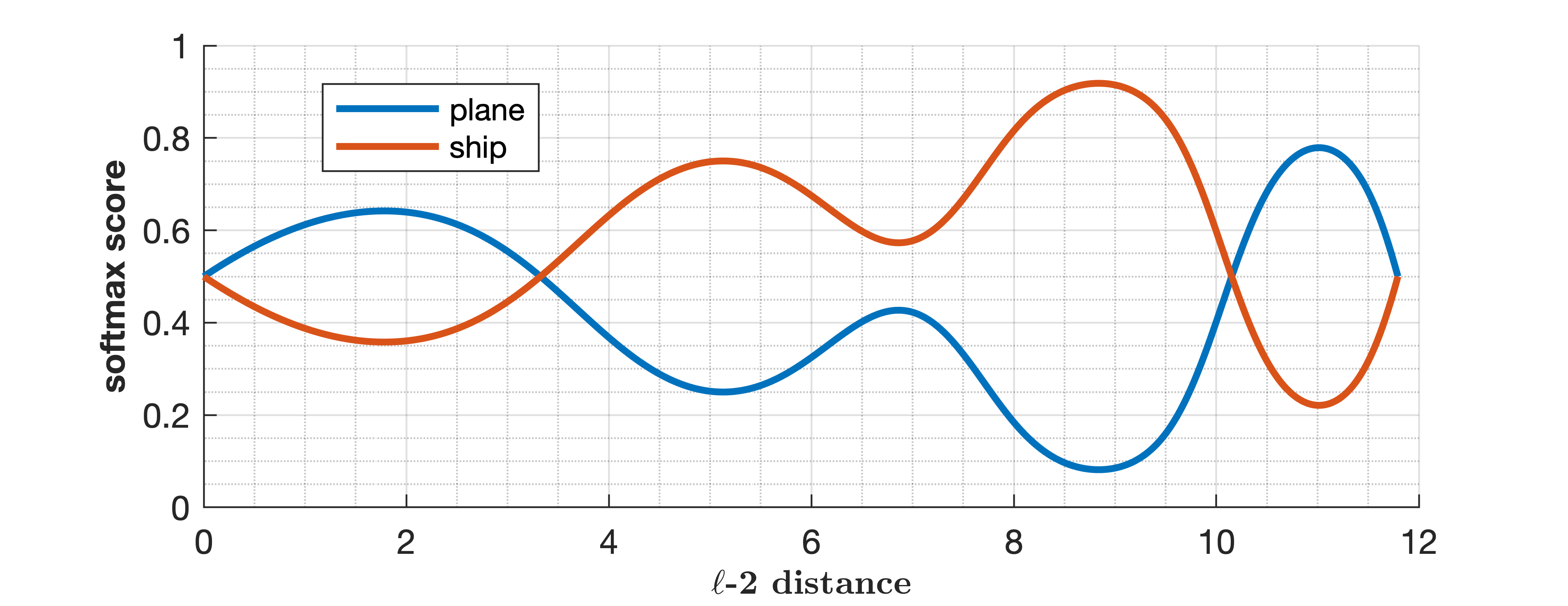} }
  \end{minipage}
  \begin{minipage}[b]{0.49\columnwidth}
\centerline{\includegraphics[width=1\columnwidth]{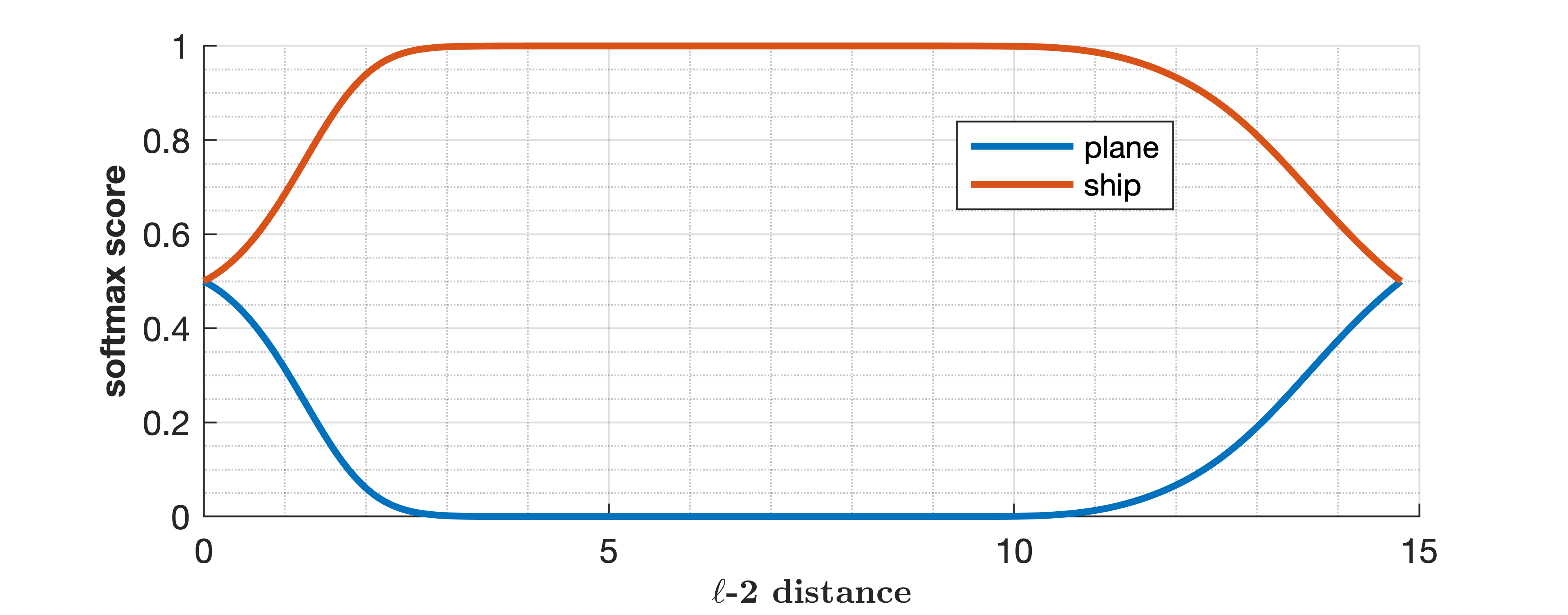} }
  \end{minipage}
  \begin{minipage}[b]{0.49\columnwidth}
\centerline{\includegraphics[width=1\columnwidth]{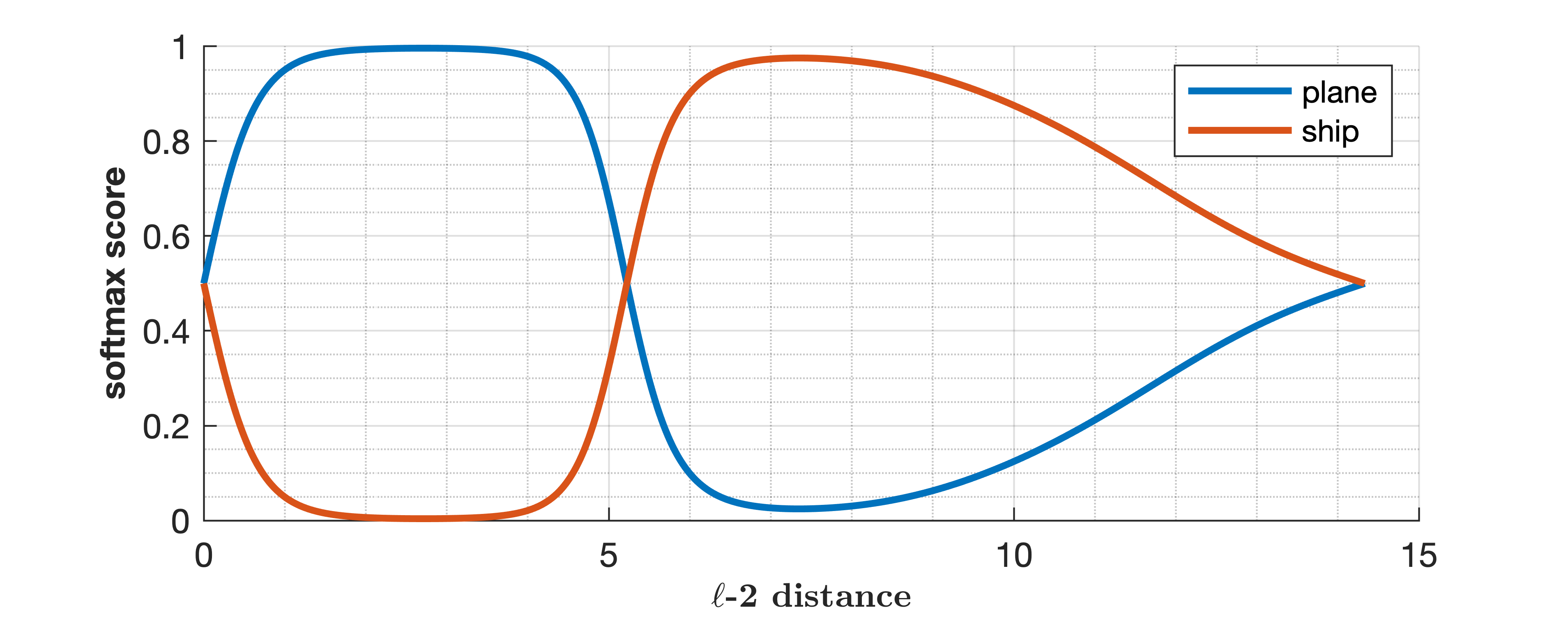} }
  \end{minipage}
  \begin{minipage}[b]{0.49\columnwidth}
\centerline{\includegraphics[width=1\columnwidth]{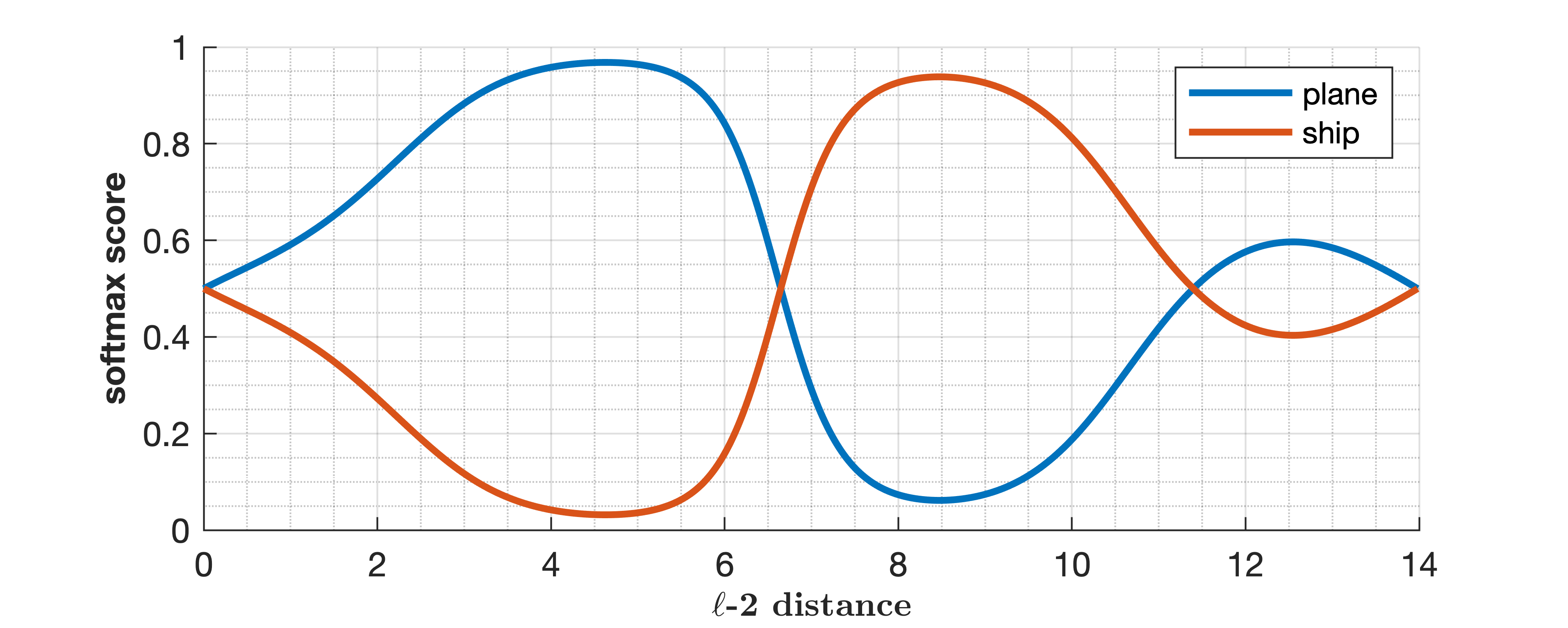} }
  \end{minipage}
\caption{Model output along the line connecting two flip points. }
\label{fig_ray3}
\end{figure}

Figure \ref{fig_ray3} considers lines connecting various pairs of flip points. If decision boundaries were linear, we would expect the red and blue curves to have $\softmax$ scores of $0.5$ all along these lines, and that is certainly not what the plots show. If decision boundaries were convex/concave, then we would expect behavior such as that in the upper right plot, but the other three plots show that the true behavior of the decision boundaries is much more complicated.

\section{Comparing with approximation methods} \label{sec_approx}

Here, we compare our calculated minimum distance to the decision boundaries with approximation methods in the literature. We also compare the direction to the closest point on decision boundary with that predicted by first order derivatives. In both comparisons we observe that relying on approximation methods may be misleading.

\begin{figure}[h]
\centering
\centerline{\includegraphics[width=.9\columnwidth]{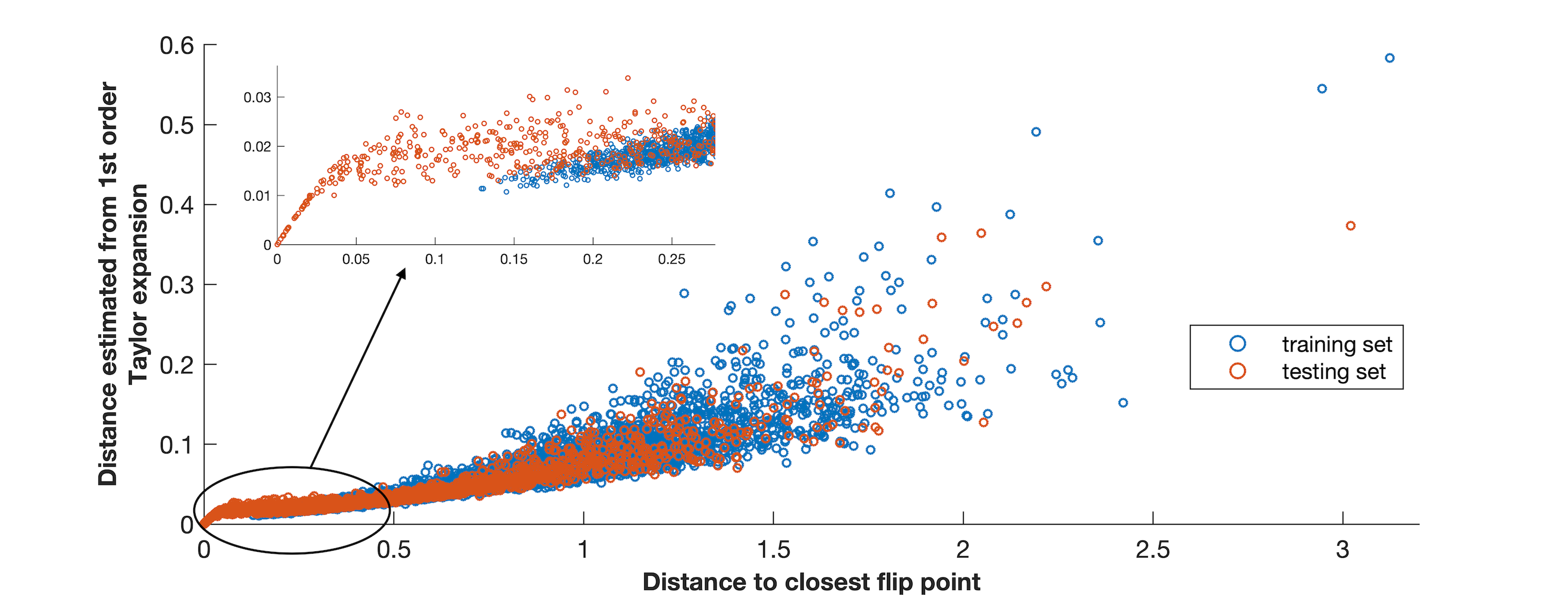} }
\caption{Using the first-order Taylor expansion for estimating the minimum distance to decision boundaries siginificantly underestimates the distance, except when points are very close to the decision boundaries (closer than 0.01).}
\label{fig_approx}
\end{figure}

\begin{figure}[h]
\centering
\centerline{\includegraphics[width=.85\columnwidth]{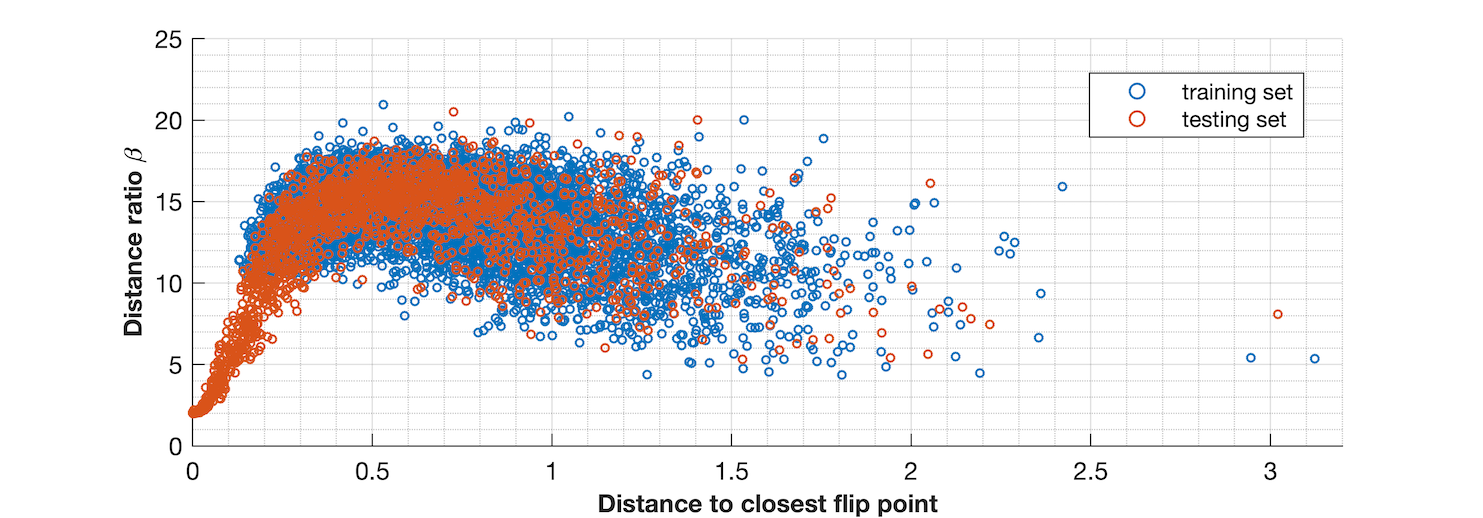} }
\caption{$\beta$ is the distance to the closest flip point divided by the distance predicted by the Taylor approximation.  Since these ratios are far from $1$, the approximation is not a reliable measure.}
\label{fig_dist_ratio}
\end{figure}

Regarding the minimum distance to the decision boundaries, \cite{elsayed2018large} suggested estimating the distance using a approximation method based on first order Taylor expansion, building on other suggestions for linear approximation of the distance, e.g., \cite{matyasko2017margin} and \cite{hein2017formal}.
The approximation method of \cite{elsayed2018large} has also been used by \cite{marginbased2019} to study the generalization error of models. 
Figure \ref{fig_approx} shows the distances computed using their approximation method versus the actual distances we have computed using flip points. For distances less than $0.01$, the Taylor approximation underestimates the distance by about a factor of $2$. For larger distances, the Taylor approximation underestimates by as much as a factor of 20 or more, as shown in Figure \ref{fig_dist_ratio}.
%is It can be seen that the first-order Taylor approximation significantly underestimates the distance to the decision boundaries, except for images that are very close to the decision boundaries (the linear section of the plot for distances less than 0.01). 
\bchange{
We note that their approximation method estimates distance to decision boundaries without finding actual points on the decision boundaries. We find that the estimated distances are generally underestimate\cchange{s} of both the true distance in that direction and the true distance to a flip point.
}
%We note that their approximation method only seeks an approximate distance to decision boundaries and they do not find actual points on the decision boundaries. Hence, their underestimation of distance does not mean that approximation has found closer points on the decision boundaries. In fact, when we investigate the direction of first order derivatives, we see that the distances to decision boundaries along that direction is much larger than the result of approximation, even larger than the distances to the closest flip points.

\begin{figure}[h]
\centering
\centerline{\includegraphics[width=.75\columnwidth]{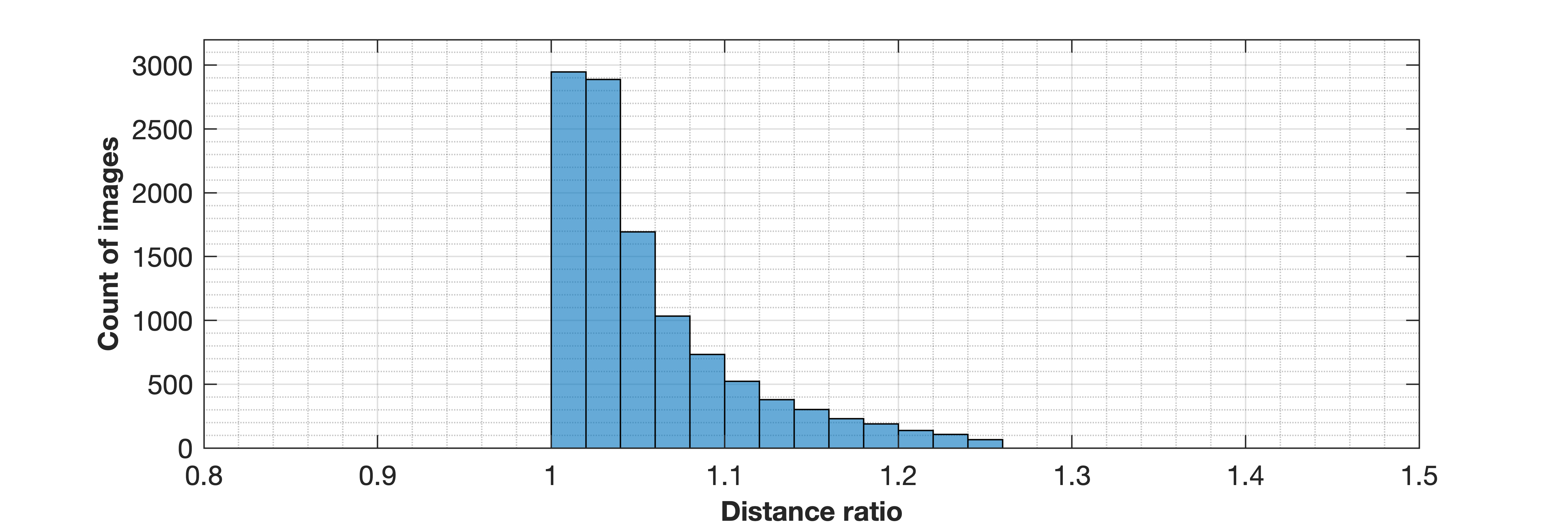} }
\caption{Finding the flip point along the direction indicated by the first-order Taylor expansion often gives an accurate estimate of distance to the decision boundary. The horizontal axis is the ratio of the distance to the decision boundary along the Taylor direction to the distance to the closest flip point.}
\label{fig_taylor}
\end{figure}

Figure \ref{fig_taylor} illustrates the \underline{distance} to the decision boundary along the direction defined by the Taylor series approximation, compared to the distance to the closest flip point. Finding the Taylor direction and then finding the intersection with the decision boundary (a one-dimensional optimization problem) are both very inexpensive operations, and our results indicate that \bchange{this approach usually gives a good approximation to the closest flip point distance (average of 1.06 times the true distance), but for 7\% of the data, a step in that direction goes outside the feasible set of images before passing through a decision boundary. This indicates that it might not be wise to limit the search to the direction of first-order derivatives. \bchange{And if we limit the search to the direction of first-order derivatives (or any other direction), it would be most reliable to search along that direction for a flip point rather than just estimating the distance.} Based on the results in Figures \ref{fig_dist_ratio} and \ref{fig_taylor}, we can conclude that the distance obtained by searching the direction of first-order derivatives can be considered a much better approximation method compared to the distance obtained by the first-order Taylor series approximation method.}

%Figure \ref{fig_dist} shows the histogram of distances to the decision boundaries, where the distribution of distance for exact points follows almost a normal distribution, while most of the approximated distances are very close to the decision boundaries. Figure \ref{fig_dist_ratio} shows the ratio of computed distances to the approximated distances ($\beta$), illustrating that for most images, the ratio can be considered almost a random number in the range of 5 to 20. Similar traits also hold for the MNIST dataset.

%\begin{figure}[h!]
%\centering
%\centerline{\includegraphics[width=.69\columnwidth]{fig_dist.png} }
%\caption{Distribution of distance to closest flip point among the images in the training and testing sets for restricted CIFAR-10 (ships and planes).}
%\label{fig_dist}
%\end{figure}

\begin{figure}[h]
\centering
\centerline{\includegraphics[width=.85\columnwidth]{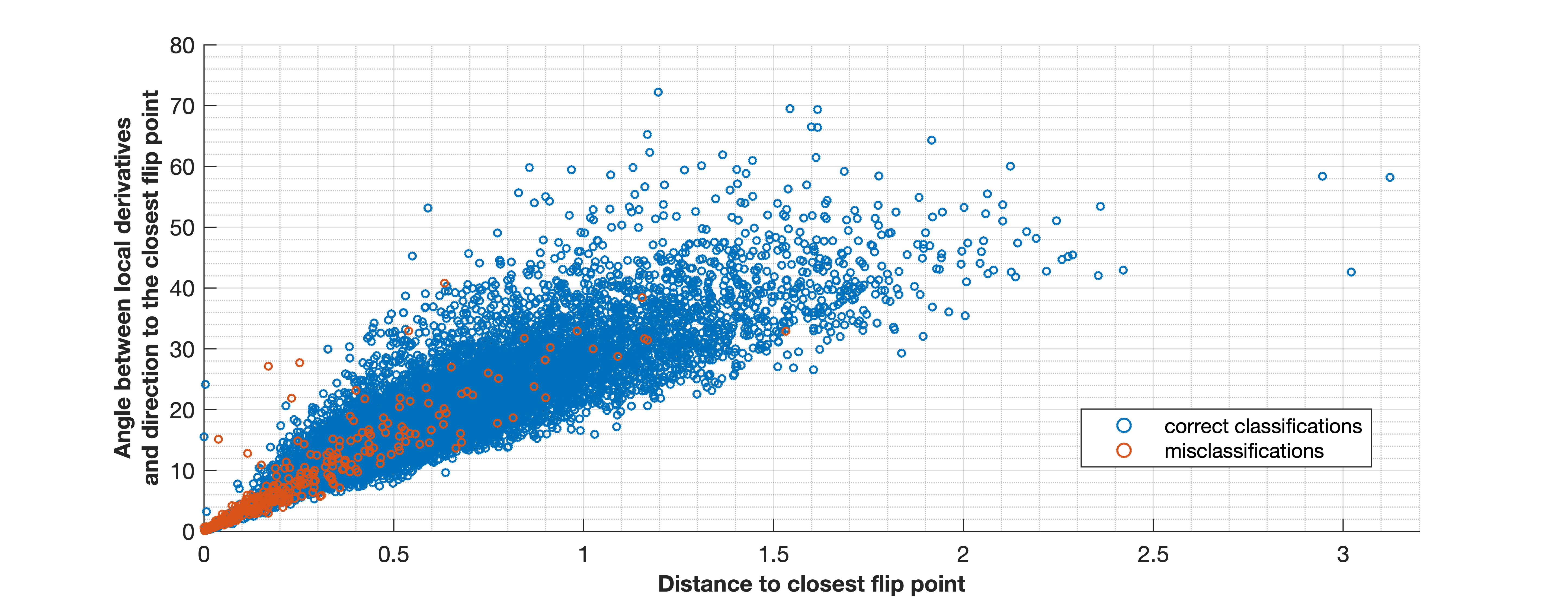} }
\caption{Angle between direction of first-order Taylor approximation and direction to closest flip points.
These angles are far from $0$, indicating that Taylor approximation gives misleading results.}
\label{fig_angle}
\end{figure}

Unfortunately, the Taylor \underline{direction} itself is not so reliable. We look at the angle between the direction defined by the Taylor approximation and that defined by the calculated closest flip point.
%Additionally, by computing the angle between the vectors, we compare the direction of first order derivatives and the direction to the closest flip point, for each image. This angle shows whether the derivatives of output of network for the image, actually point toward the closest flip points or not. 
Large angle between the two directions means the derivative does not point near the closest point on decision boundary. Figure \ref{fig_angle} shows the distribution of the angles (in degrees) vs the distance to the closest flip points. This clearly shows that the farther an image is from the decision boundary, the larger the angle tends to be. In Figure \ref{fig_angle}, we observe that the lower bound for the angles linearly increases with the distance. 

\achange{All these observations show that the simplifying assumptions used by \cite{elsayed2018large} and \cite{marginbased2019} can be unreliable, and signify the importance of verification for such simplifying assumptions,} \cchange{whenever used for models with nonlinear activation functions.}

%The comparison we presented here, is based on the closest flip points we have computed by solving a nonlinear non-convex optimization problem, therefore, we cannot be completely sure that the closest flip points we have found are actually the closest points on the decision boundaries. Nevertheless, based on the results we presented here, it is clear that first order Taylor expansion has major limitations in estimating the distance to the closest flip points.

\section{Shape and connectedness of decision regions}

%To investigate the surface of decision boundaries, we find 201 distinct flip points on a contiguous decision boundary for the restricted CIFAR-10 model. This allows us to estimate the curvature of the surface of decision boundary. We compute the flip points such that one of them is the closest flip point to one of the images and the rest of them are within 0.1 distance of all the other 200 flip points. We also ensure the direct path between all pairs of flip points is on one side of the decision boundary. This arrangement of flip points ensures that the curvature we obtain corresponds to a small region on the decision boundary contained in a ball of diameter 0.1, and the direction of the bend of the surface is the same for that region.

%We perform this for 100 randomly chosen images in the training and testing sets, and see that the curvature is far from linear for all of them. Figure \ref{} is the histogram of estimated curvatures for those images. This reveals the assumptions about local linearity of decision boundaries may be unreliable, even within very small regions.

%\begin{figure}[h]
%\centering
%\centerline{\includegraphics[width=.7\columnwidth]{fig_clust_eig.png} }
%\caption{The number of large eigenvalues of an adjacency matrix reveals the number of major clusters within a graph. Clearly our testing set consists of very few clusters.}
%\label{fig_clust_eig}
%\end{figure}

We consider all images of correctly-classified ships in the testing set, and investigate the lines (in image space) connecting each pair of images. 
89\% of those lines stay within the ``ship" class for the model, while 11\% do not. The least-connected ship is connected to  220 other ships by lines that do not exit the ``ship" region, and there are paths (some using multiple lines) that connect every pair of ships without exiting the ``ship" region. This indicates that
the ``ship" region is star-shaped, providing another reason why linear approximations to decision boundaries are inadequate.
These observations also hold for images in the training set. \bchange{Therefore, the trained network has formed a connected sub-region (in the domain) that defines the ``ship" region. This result aligns with the observations reported by \cite{fawzi2018empirical} that classification regions created by \cchange{a deep neural network can be connected and a single large region may} contain all points of the same label. \cite{fawzi2018empirical}, however, did not investigate the output of network along direct paths between images of the same class.}

We performed our analysis by building the adjacency matrix of directly connected images. Performing spectral clustering \citep{von2007tutorial} on the graph and the Laplacian derived from the adjacency matrix that incorporates the distance between images, may provide additional insights.

\section{Adversarial examples and decision boundaries}

In most recent studies about adversarial examples, the inputs with adversarial label are obtained by minimizing the loss function of the neural network for that label, subject to a distance constraint \citep{tsipras2018robustness,ilyas2019adversarial}. The distance constraint keeps the adversarial image close to the original image. There are limitations to this approach, as we explain via examples. 
%As we showed earlier, the distance to the decision boundaries vary among the images in the training set and testing set.However, when we compute the closest flip point on the decision boundary of the model, a point arbitrarily close to that flip point will have the opposite label.

\begin{figure}[h]
\centering
  \begin{minipage}[c]{0.19\columnwidth}
\centerline{\includegraphics[width=1\columnwidth]{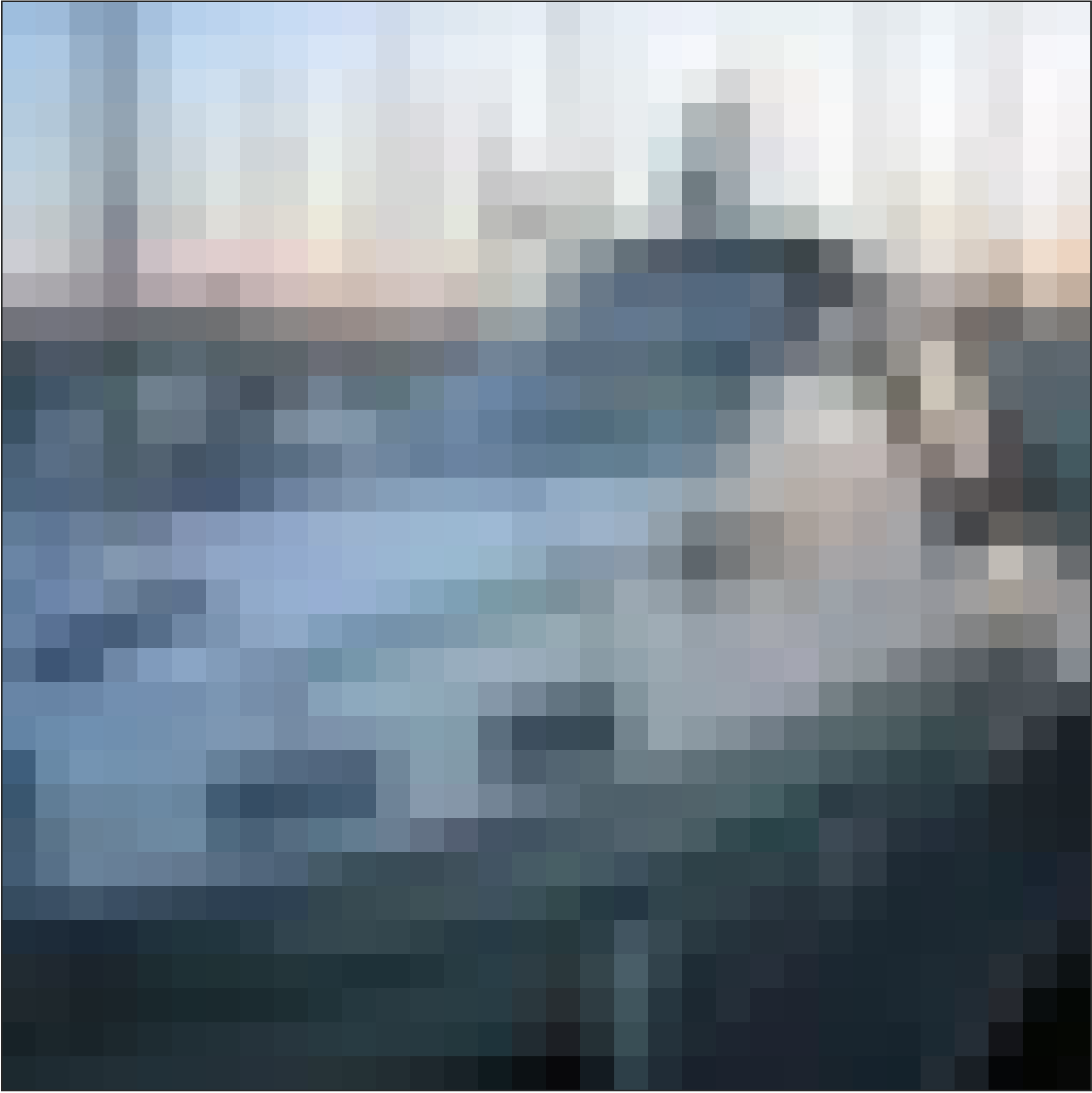} }
  \end{minipage}
  \begin{minipage}[c]{0.7\columnwidth}
\centerline{\includegraphics[width=1\columnwidth]{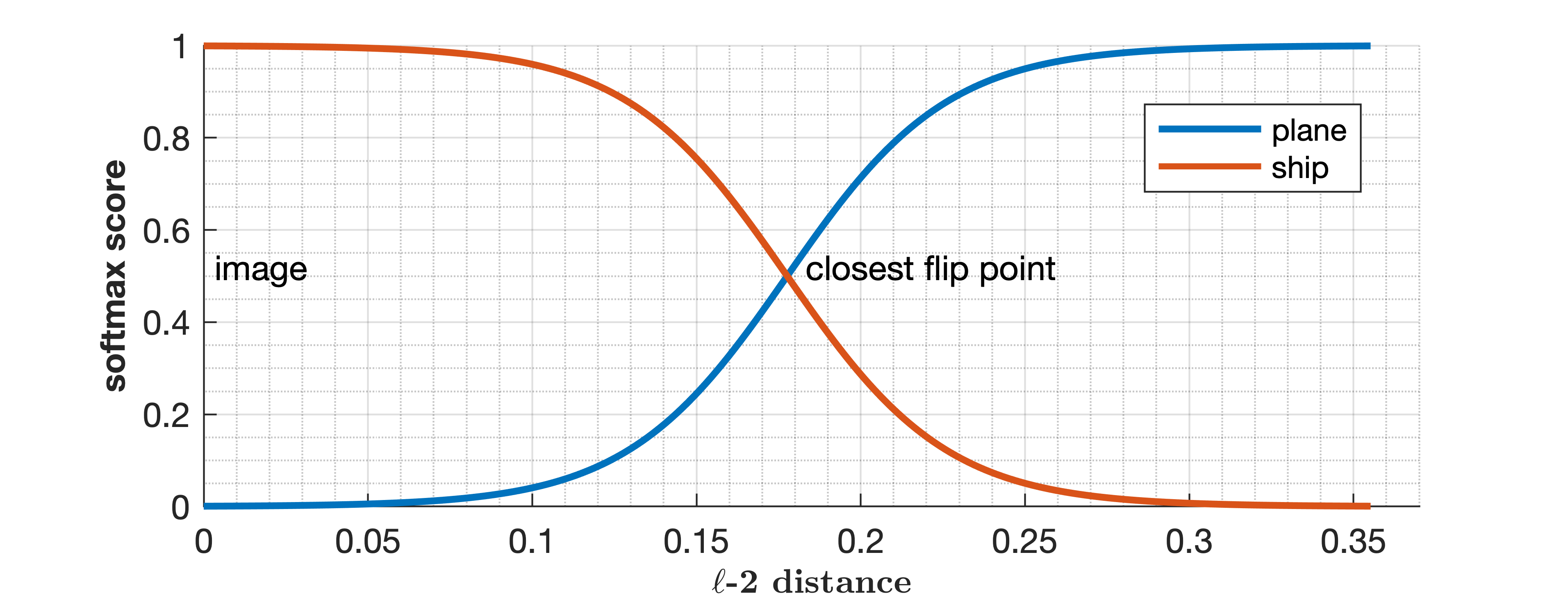} }
  \end{minipage}
\caption{Finding the closest flip point reveals the least changes that would lead to an adversarial label for the image.}
\label{fig_adv1}
\end{figure}

\begin{figure}[h]
\centering
\centerline{\includegraphics[width=.7\columnwidth]{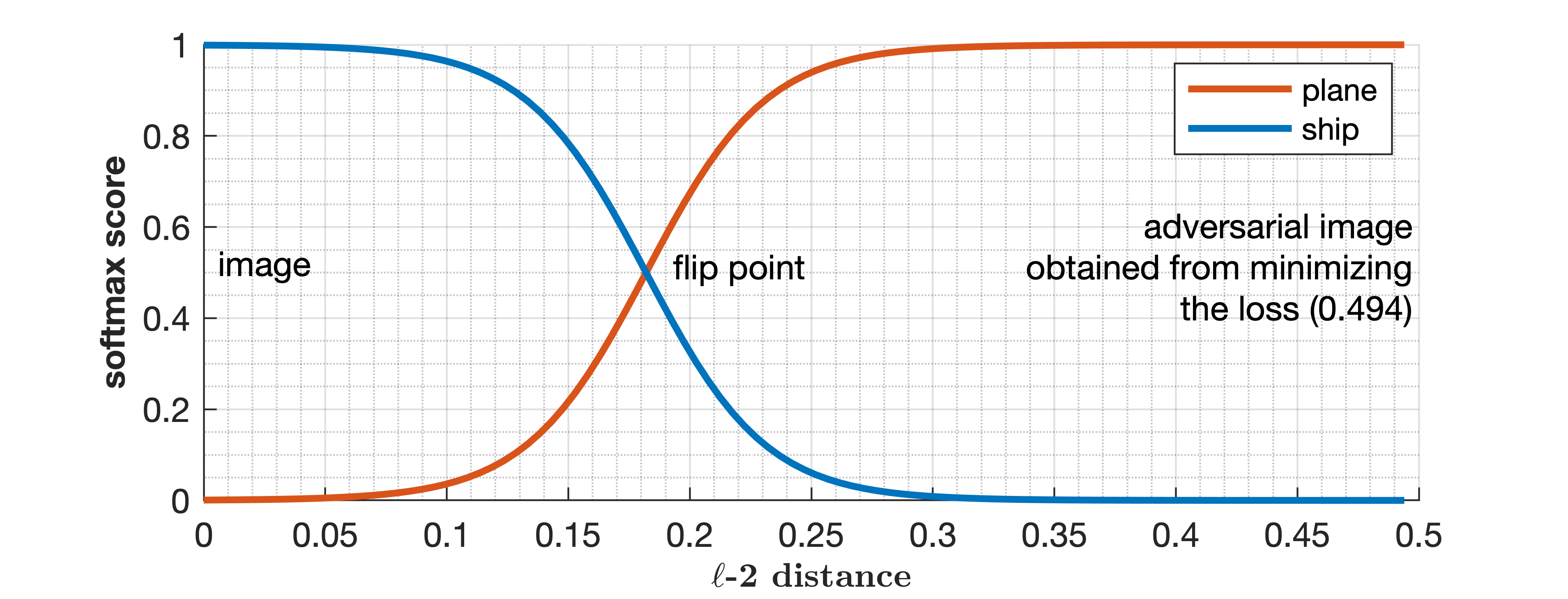} }
\caption{Minimizing the loss function subject to a distance constraint may find adversarial examples far from the original image. }
\label{fig_adv2}
\end{figure}

Consider the  image on the left in Figure \ref{fig_adv1}. Figure \ref{fig_adv1} (right) shows the value of the $\softmax$ score on the line from this image to its closest flip point and  beyond.
We compare this with the result of  minimizing the loss function of the model for the adversarial label ``plane", subject to $\ell_2$ distance constraint of 0.5, as suggested by \cite{ilyas2019adversarial}. The adversarial image obtained by this method is much further away, a distance of 0.494 instead of a distance of  0.178 for the closest flip point that we found. So their calculation underestimates the vulnerability of the model.
It is also interesting that the line between the image and the adversarial image found by their method crosses a flip point at a distance much less than 0.494, as shown in Figure \ref{fig_adv2}, yielding a much better assessment of the vulnerability of the model.
%adversarial images much closer than and the direct path between them is shown in Figure %\ref{fig_adv2}. As we can see in both Figures \ref{fig_adv1} and \ref{fig_adv2}, there are adversarial images much closer to the original image, that have $\softmax$ score close to 1 (images as close as 0.3). Moreover, there are images arbitrary close to the closest flip point that have $\softmax$ score greater than half for the adversarial label. However, finding the adversarial example with the distance constraint, may miss all of that information, and may not find the weakest vulnerability of the model.

\begin{figure}[h]
\centering
  \begin{minipage}[c]{0.19\columnwidth}
\centerline{\includegraphics[width=1\columnwidth]{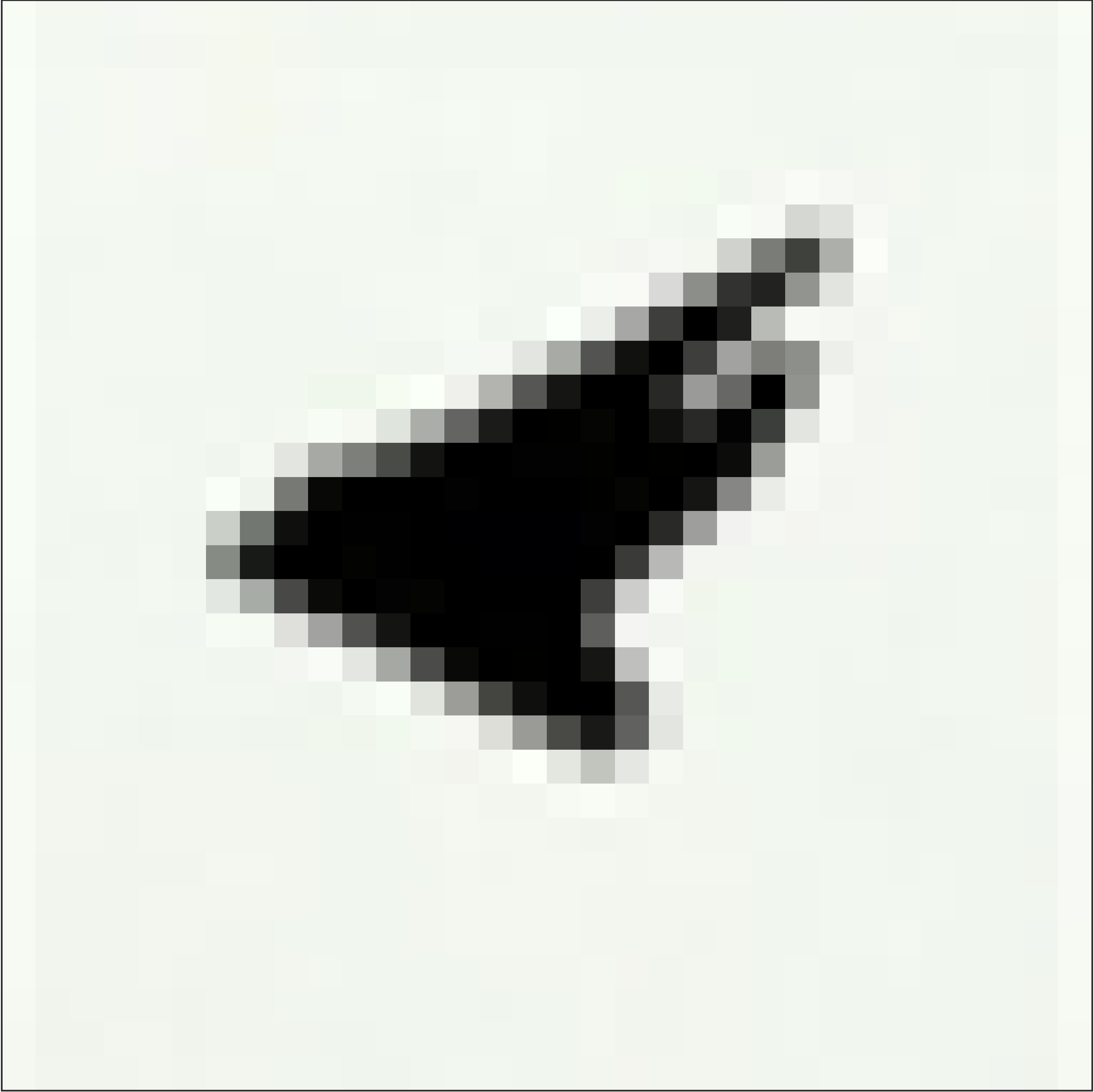} }
  \end{minipage}
  \begin{minipage}[c]{0.7\columnwidth}
\centerline{\includegraphics[width=1\columnwidth]{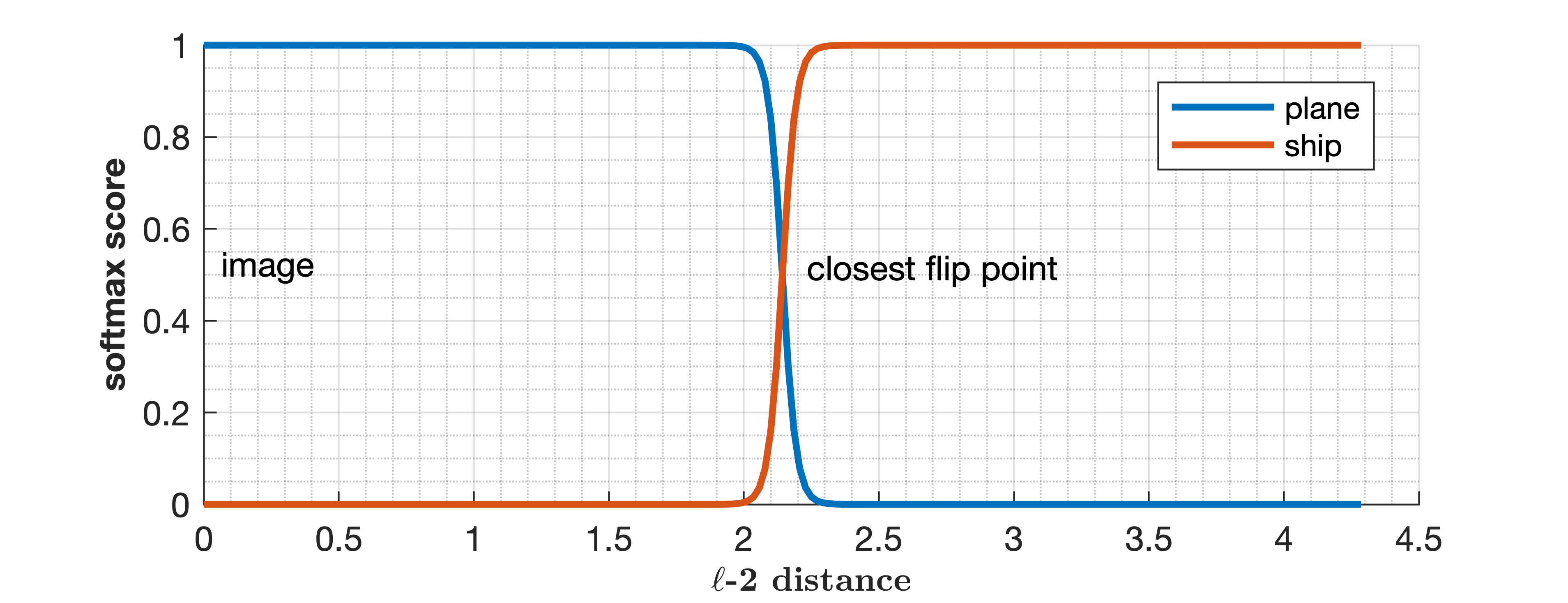} }
  \end{minipage}\caption{Minimizing the loss function subject to a tight distance constraint may not have a feasible solution and would not reveal how robust the model actually is. }
\label{fig_adv3}
\end{figure}

There is another difficulty associated with minimizing the loss for an adversarial label subject to a distance constraint. As an example, consider the image in Figure \ref{fig_adv3}, which is at a  distance of  2.14 from its closest flip point. Seeking an adversarial image for this image with distance constraint 0.5 will be unsuccessful, as the optimization problem has no feasible solution. 
Finding the closest flip point yields much better information about robustness.
%But, this is not completely informative as to how robust the model is. Luckily, the closest flip point can accurately reveal the measure of robustness.

We also measure the angle between the direction to the closest flip point, and the direction to the adversarial example found by minimizing the loss function. For the image in Figure \ref{fig_adv2}, the angle is 12.7 degrees.
%, and for the image in Figure \ref{fig_adv3}, the angle is 67.7 (when an adversarial example at a larger distance is allowed).

The cost of finding a flip point is comparable to the cost of minimizing the loss function, and it provides much better information. %\achange{Note that minimizing the loss function is also non-convex, same as the optimization problem we have defined. So, non-convexity is not a disadvantage for our method.}

\achange{The distance constraint used by \cite{ilyas2019adversarial} can be viewed as a ball around the input. We showed that choosing the size of that ball can be challenging. If the size of ball is too small, their optimization problem becomes infeasible. If the size of ball is large, they do not find the adversarial example closest to the input. Their problem is non-convex, like ours. The examples above demonstrated that our approach finds a closer adversarial example compared to their approach. The computation times for our method and theirs are quite similar.}

\bchange{
Moreover, Figure \ref{fig_adv4} shows that the closest distance to the decision boundary can have a large variation among the images in a dataset. Therefore, tuning the distance constraint for one image may not be insightful for most of the other images in a dataset.

\begin{figure}[h]
\centering
\centerline{\includegraphics[width=.75\columnwidth]{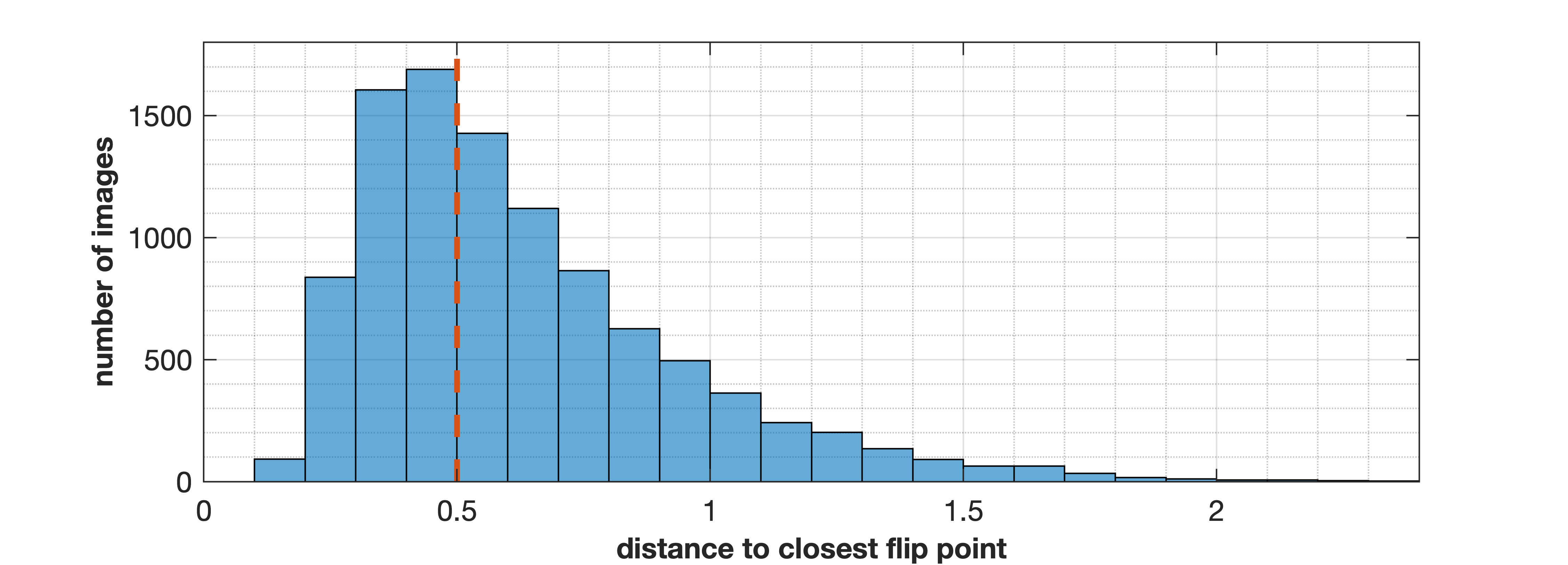} }
%  \vspace{.2in}
\caption{Distance to the closest flip point has large variation among images in the training set, which shows that a single distance constraint would not be able to reveal the vulnerabilities of a model for all images. For example, a distance constraint of 0.5 cannot yield an adversarial example for the large fraction of images that are farther than 0.5 from the decision boundaries. It also would not reveal the weakest vulnerabilities for images which are much closer than 0.5 to the decision boundaries.}
\label{fig_adv4}
\end{figure}
%\linespread{2}
}
%\bchange{But, guessing the value of distance constraint for each image is not practical either. }

These observations would still hold for networks trained on the pixels rather than wavelet coefficients.

Regarding the reason for excessive vulnerability of trained neural networks towards adversarial examples \citep{goodfellow2014explaining}, there are studies that speculate about decision boundaries. For example, \cite{tanay2016boundary} argue that ``adversarial examples exist when the classification boundary lies close to the submanifold of sampled data", but their analysis is limited to linear classifiers. \cite{shamir2019simple} also explain the adversarial examples via geometric structure of partitions defined by decision boundaries; however, they do not consider the actual distance to the decision boundaries, nor the feasibility of changes in the input space, and their analysis is focused on linear decision boundaries.

The analysis provided in this paper show\cchange{s} that studies focused on adversarial examples can benefit from using the closest flip points and from direct investigation of decision boundaries, for measuring and understanding the vulnerabilities, and for making the models more robust.

\section{Conclusions}

We showed the complexities of decision regions of a model can make linear approximation methods quite unreliable\bchange{, when nonlinear activation functions are used for the neurons}. Instead, we used flip points to provide improved estimates of distance and direction of data points to decision boundaries. These estimates can provide measures of confidence in classifications, explain the smallest change in features that change the decision, and generate adversarial examples. \cchange{Closest} flip points are computed by solving a non-convex optimization problem, but the cost of this is comparable to methods used to compute an adversarial point that may be much further away. \cchange{The closest flip point along a particular direction can be easily computed by a bisection algorithm.} Our example involved only two classes and continuous input data, but we have also implemented our method for problems with multiple classes and discrete features.

%\newpage

\bibliography{Refs}
\bibliographystyle{plainnat}

\clearpage
\setcounter{table}{0}
\renewcommand{\thetable}{A\arabic{table}}

\section*{Appendix A: Information about neural network used in our numerical examples}

Here, we provide more information about the model we have trained and used in the paper. 
Our model is a fully connected feed-forward neural network with 12 hidden layers. The inputs to the model are 200 wavelet coefficients for any image, as explained earlier. The number of neurons for each layer are shown in Table \ref{table_nodes}.

\begin{table}[h]
\setlength{\tabcolsep}{5pt}
\caption{Number of nodes in neural network used for each data set.}
\label{table_nodes}
\vskip 0.15in
\begin{center}
\begin{small}
\begin{sc}
\begin{tabular}{  r | r }
\toprule
Data set & Restricted CIFAR-10   \\
\midrule
Input layer & 200  \\
\midrule
Layer 1 & 700  \\
\midrule
Layer 2 & 600 \\
\midrule
Layer 3 & 510 \\
\midrule
Layer 4 & 440  \\
\midrule
Layer 5 & 375  \\
\midrule
Layer 6 & 325  \\
\midrule
Layer 7 & 285  \\
\midrule
Layer 8 & 250 \\
\midrule
Layer 9 & 215 \\
\midrule
Layer 10 & 160 \\
\midrule
Layer 11 & 100 \\
\midrule
Layer 12 & 40  \\
\midrule
Output layer & 2  \\
\bottomrule
\end{tabular}
\end{sc}
\end{small}
\end{center}
\vskip -0.1in
\end{table}

The activation function we have used in the nodes is the error function
\begin{center}
$activation(y|\sigma) = \erf(\frac{y}{\sigma}) = \frac{1}{\sqrt{\pi}} {\bigintss}_{\hspace{-3pt}-\frac{y}{\sigma}}^{+\frac{y}{\sigma}} {e}^{-t^{2}} { d}t,$
\end{center}
where $y$ is the result of applying the weights and bias to the node's inputs. The tuning parameter $\sigma$ is constant among the nodes on each layer and is optimized during the training process.

We have used $\softmax$ on the output layer, and cross entropy for the loss function.

%\clearpage
%\setcounter{table}{0}
%\renewcommand{\thetable}{B\arabic{table}}
%
%\section*{Appendix B: Lipschitz continuity of trained neural network}
%Here, we investigate the Lipschitz continuity of the trained neural network we have used in our numerical experiments.
%

\end{document}